\documentclass[twoside,11pt]{article}

\usepackage{jmlr2e}

\usepackage{amsmath, amssymb, amsfonts}

\usepackage{xcolor}

\usepackage{algorithm}
\usepackage{algpseudocode}

\usepackage{lastpage}

\newcommand{\Emph}[1]{\textbf{#1}}
\newcommand{\Integer}{\mathbb{Z}}
\newcommand{\Real}{\mathbb{R}}
\newcommand{\SymbolSet}{\Sigma}
\newcommand{\Symbola}{w} 
\newcommand{\Time}{t}
\newcommand{\String}{s}
\newcommand{\StringRV}{S}
\newcommand{\Output}{y}
\newcommand{\OutputRV}{Y}
\newcommand{\StringLen}{n}
\newcommand{\IDatum}{i}
\newcommand{\DataLen}{m}
\newcommand{\Trainer}{\mathfrak{A}}
\newcommand{\NewTerm}[1]{\textbf{\emph{#1}}}
\newcommand{\LetterInstance}[1]{\textrm{`\texttt{#1}'}}
\newcommand{\StringInstance}[1]{``\texttt{#1}''}
\newcommand{\LM}{h}
\newcommand{\LMSet}{\mathcal{H}}
\newcommand{\AcceptableMap}{F_{0}}
\newcommand{\AcceptableFunc}{f_{0}}
\newcommand{\PMF}{P}
\DeclareMathOperator{\CDF}{CDF}
\newcommand{\Measure}{\mu}
\newcommand{\MeasureSet}{\mathcal{P}}
\newcommand{\MeasureSetOn}{\Delta}

\DeclareMathOperator{\Probability}{Pr}
\newcommand{\HallucinationSymbol}{\mathrm{H}}
\newcommand{\TrainSymbol}{\mathrm{T}}
\DeclareMathOperator{\Length}{len}
\DeclareMathOperator{\HallucinationProbability}{HP}

\newcommand{\TrainingDataRV}{T}
\DeclareMathOperator*{\argmin}{argmin}

\newcommand{\IString}{j}
\newcommand{\NStrings}{k}
\newcommand{\PrMass}{p}
\newcommand{\HallucinationEvent}{U}
\newcommand{\StringMap}{\psi}
\newcommand{\HallucinationSet}{\tilde{\mathcal{I}}}
\newcommand{\Domain}{\mathcal{X}}
\newcommand{\Codomain}{\mathcal{Y}}
\newcommand{\InputRV}{X}
\newcommand{\Learner}{\mathfrak{A}}
\newcommand{\Hypothesis}{h}

\DeclareMathOperator{\Uniform}{Uni}
\newcommand{\LowerBound}{\lambda}
\newcommand{\NFuncs}{Q}
\newcommand{\IFunc}{q}
\newcommand{\Func}{f}

\newcommand{\Expect}{\mathbb{E}}
\newcommand{\IDataset}{d}
\newcommand{\NDatasets}{D}
\newcommand{\Input}{x}
\newcommand{\InputSeq}{\boldsymbol{x}}

\newcommand{\NInputs}{n}

\newcommand{\NLabels}{p}
\newcommand{\InputSeqRV}{\boldsymbol{X}}

\newcommand{\underlineDomain}{\underline{\Domain}}
\newcommand{\underlineCodomain}{\underline{\Codomain}}
\newcommand{\HPExpect}{\mu}

\title{%
Hallucinations are inevitable \\ but can be made statistically negligible. 
}

\author{%
\name Atsushi Suzuki \email suzuki@hku.hk \\
\addr Department of Mathematics\\
The University of Hong Kong\\
Pokfulam Road, Hong Kong
\AND
\name Yulan He \email yulan.he@kcl.ac.uk \\
\addr Department of Informatics\\
King's College London\\
30 Aldwych, London, WC2B 4BG, United Kingdom
\AND
\name Feng Tian \email feng.tian978@dukekunshan.edu.cn \\
\addr Division of Natural and Applied Sciences\\
Duke Kunshan University\\
No. 8 Duke Avenue, Kunshan City, Jiangsu Province, 215316, China
\AND
\name Zhongyuan Wang \email wzy\_hope@163.com \\
\addr School of Computer Science\\
Wuhan University\\
No. 299 Bayi Road, Wuchang District, Wuhan City, Hubei Province, China
}

\editor{My editor}

\jmlrheading{23}{2022}{1-\pageref{LastPage}}{1/21; Revised 5/22}{9/22}{21-0000}{Atsushi Suzuki, Yulan He, Feng Tian, and Zhongyuan Wang}

\ShortHeadings{Hallucinations are inevitable but can be made statistically negligible}{Suzuki, He, Tian, and Wang}
\firstpageno{1}

\begin{document}

\maketitle

\begin{abstract}
Hallucinations, a phenomenon where a language model (LM) generates nonfactual content, pose a significant challenge to the practical deployment of LMs. While many empirical methods have been proposed to mitigate hallucinations, recent studies established a computability-theoretic result showing that any LM will inevitably generate hallucinations on an infinite set of inputs, regardless of the quality and quantity of training datasets and the choice of the language model architecture and training and inference algorithms. Although the computability-theoretic result may seem pessimistic, its significance in practical viewpoints has remained unclear. This paper claims that those "innate" inevitability results from computability theory and diagonal argument, in principle, cannot explain practical issues of LLMs. We demonstrate this claim by presenting a positive theoretical result from a probabilistic perspective. Specifically, we prove that hallucinations can be made statistically negligible, provided that the quality and quantity of the training data are sufficient. Interestingly, our positive result coexists with the computability-theoretic result, implying that while hallucinations on an infinite set of inputs cannot be entirely eliminated, their probability can always be reduced by improving algorithms and training data. By evaluating the two seemingly contradictory results through the lens of information theory, we argue that our probability-theoretic positive result better reflects practical considerations than the computability-theoretic negative result. 
\end{abstract}


\section{Introduction}

A language model (LM), in a broad sense, is a computer program to solve a task whose input and/or output are natural language sentences.
Typically, both the input and output of the task are formulated as natural language sentences.
For example, in scenarios like translation or chatbots, the task is to receive natural language sentence input that users type or chat and generate a natural language sentence output that meets the users' desire described in the input sentences.
Early approaches relied on rule-based pattern matching, e.g., ELIZA \citep{weizenbaum1966eliza}, PARRY \citep{colby1971artificial}, ALICE \citep{wallace2009anatomy}, etc. or statistical language models based on Markov theories, e.g., \citep{kuhn1990cache,hiemstra1998linguistically,chen1999empirical}.
However, the introduction of artificial neural networks in LMs, pioneered by, e.g., \citep{rumelhart1986learning,elman1990finding,mahoney2000fast,bengio2000neural}, has led to a paradigm shift over the past two decades, as advances in techniques and hardware have enabled large-scale neural models. 
The techniques supporting the success include effective neural network architectures, e.g., long short term memory \citep{hochreiter1997long,gers2000learning}, the encoder-decoder model \citep{cho2014learning}, the attention architecture \citep{bahdanau2014neural}, Transformer \citep{vaswani2017attention}, etc., pretraining strategies, e.g., BERT \citep{devlin2019bert}, and learning strategies human feedback, e.g., \citep{ouyang2022training}.
Those large-scale neural language models, often simply called large language models (LLMs), have impacted academia and society, represented by the launch of powerful chatbots, e.g., ChatGPT \citep{radford2019language,brown2020language,achiam2023gpt}, Gemini \citep{team2023gemini,team2024gemini}, LLaMA \citep{touvron2023llama,touvron2023llama2,dubey2024llama}, Claude \citep{anthropic2024claude}, Qwen \citep{bai2023qwen,yang2024qwen2,yang2024qwen25}, DeepSeek \citep{liu2024deepseek,guo2025deepseek}, etc., as well as success in the fields of machine translation \citep{wu2016google}, search engine \citep{microsoft2024new}, recommendation systems \citep{li2023gpt4rec,gao2023chat}.
For more details of language models, refer to, e.g., \citep{zhao2023survey,minaee2024large,dam2024complete}. 

As LLMs have impacted society, \NewTerm{hallucinations} have been identified as crucial issues, complicating their practical deployment in applications \citep{huang2023survey}. 
Here, hallucinations are defined as a phenomenon where a LM generates nonfactual content \citep{huang2023survey} or content nonsensical or unfaithful to the provided source \citep{ji2023survey}.
The root causes of hallucinations have generally been categorized \citep{ji2023survey,huang2023survey} into data, training, and inference, and
many empirical methods have been proposed to mitigate hallucinations, e.g., by exploiting knowledge bases \citep{shuster2021retrieval,zhao2023verify}, refining the requirement on LMs \citep{wei2022chain,dhuliawala2023chain}, or applying information theoretic methods \citep{farquhar2024detecting}. 
However, based on computability theory, a recent study \citep{xu2024hallucination} has proved using diagonal argument that in a certain ground truth setting, any LM \textemdash regardless of its training and inference algorithms or training dataset employed \textemdash will inevitably produce hallucinations on an infinite set of input strings. 
Another study \citep{banerjee2024llms} construct a concrete example of an inevitable hallucination instance by reducing the problem to the halting problem, which is also essentially based on diagonal argument.
This theoretical result may seem fatally pessimistic for practitioners since hallucinations on infinite input instances sound like an insurmountable obstacle in practice.
Indeed, those results have been referred to as fundamental limitations of LLMs not only in academia but also in the general public, e.g., articles in an online encyclopedia \citep{wiki2025hallucination} and written by a journalist \citep{jones2025ai},
being referred to as a ground to state that we need to "live with them (hallucinations)" \citep{banerjee2024llms} or hallucinations "can't be stopped" \citep{jones2025ai}.

However, generally speaking, the implications of computability-theoretic theorems based on diagonal argument need to be carefully discussed from practical viewpoints.
For example, although there exist uncountably infinite non-computable mathematical functions, computers have been significantly useful in computing plenty of practical functions.
Hence, it is crucial to know what the theoretical result by \citep{xu2024hallucination} actually implies from more practical viewpoints.
In other words, our question is the following: \Emph{Can the inevitability of hallucinations have practical implications?}
Since the core of those discussions is in non-computable functions, no numerical simulations can rebut them. Hence, theoretical discussions are necessary.

This paper claims that \Emph{those "innate" inevitability results from computability theory and diagonal argument, in principle, cannot explain practical issues of LLMs}, by presenting a contrastive, positive theoretical result from a probabilistic perspective on a problem setting compatible with the previous work \citep{xu2024hallucination}. 
Specifically, we show that we can reduce the probability of hallucinations arbitrarily close to zero, provided that the training data is of sufficient quality and quantity and certain training and inference algorithms are employed.
In other words, even if LLMs fail, \Emph{we should ascribe the failure to the algorithm and the quality or quantity of the dataset, not to the "innate" inevitability of hallucinations proved by diagonal arguments.}
Crucially, our positive result mathematically coexists with the negative result of \citep{xu2024hallucination} under a wide range of settings.
We also solve the paradox behind the coexistence, recalling that an infinite set equipped with a probability measure can have an infinite subset with an arbitrarily small probability.
In other words, even though we cannot avoid hallucinates on infinite input instances, it is still possible to reduce the \Emph{probability} of hallucinations arbitrarily close to zero by improving training data and training and inference algorithms.
The practical significance of hallucinations occurring only on infinite input sets with arbitrarily small probability can ultimately depend on the application domain. Still, using Shannon's coding theorem as an example, we point out that, in the field of information theory, errors occurring on subsets with arbitrarily small probabilities are often considered negligible in practice. 
Thus, based on our theoretical result, we can conclude that hallucinations are practically negligible in domains where information theory has been successfully applied without practical issues, provided the quality and quantity of training data are sufficient. In other words, where hallucinations are indeed a practical issue, the cause should be attributed to either the dataset (quality or quantity) or the algorithm, but not to an "innate" inevitability of hallucinations derived from computability theory based on the diagonal argument.

Note that we make no assumptions regarding the grammatical or semantic structure of natural language or the nature of the ground truth mapping. 
This is a significant advantage of our theorems since natural languages are not considered to satisfy mathematically convenient conditions completely.

The contributions of the paper are listed as follows:
\begin{itemize}
    \item We show in the discrete setting reflecting natural language processing that hallucinations are statistically negligible with an appropriate algorithm and the quality and quantity of training data, provided prior knowledge about the input length distribution is available.
    \item We evaluate the statistical negligibility and inevitability of hallucinations through the lens of information theory, arguing that the statistical negligibility better reflects practical considerations.
\end{itemize}

\subsection{Related work}
Our work is directly inspired by
\citep{xu2024hallucination}, stating that any language model hallucinates on infinite input instances. 
While they evaluate hallucinations on computability theory only, our work evaluates hallucinations from both computability and probability perspectives to clarify the theories' implications from practical viewpoints.
\citep{kalai2024calibrated} clarified that hallucinations are inevitable when the real-world distribution and the training data distribution are different. 
However, such situations are out of our scope since we are interested in the situation where training data is qualified, where the computability-theoretic limitation still holds.

Some work has focused on specific neural network architectures based on Transformer \citep{vaswani2017attention} in the continuous function approximation context.
For example, \citep{yun2020transformers} and \citep{zaheer2020big} have proved that Transformers are universal
approximators of continuous sequence-to-sequence functions with compact support, though they suffer from the curse of dimensionality under their assumptions.
Transformers have been known to avoid the curse of dimensionality with stronger assumptions on the function space, such as sparse boolean functions \citep{edelman2022inductive}, hierarchical compositions \citep{gurevych2022on}, and shift-invariant and piecewise smooth functions \citep{takakura2023approximation,kim2024transformers}.
However, the continuous function framework in these studies is different from the discrete set framework that \citep{xu2024hallucination} and our work consider to be modeling natural language tokens directly.
Since the computability-based limitation proved by \citep{xu2024hallucination} comes essentially from the discrete set setting, the theoretical framework must be based on the same setting with minimal assumptions so that we can compare the result with the limitation proved by \citep{xu2024hallucination}.
We also point out a similar problem setting is intensively considered by \citep{agarwal2020learnability}, while their motivation is in how the probably approximately correct (PAC) learnable changes where we restrict the hypothesis class to computable functions rather than in evaluating the probability of hallucinations caused by the computability limitation.

Technically, our theorems are straightforwardly derived from either computability theory, intensively used in \citep{xu2024hallucination}, or the classical no-free-lunch theorem in statistical learning theory in, e.g., \citep{shalev2014understanding}.
Our technical contribution lies in providing an integrated framework to discuss LMs from the two completely different theories at the same time, rather than novel proof techniques.

The remainder of the paper is organised as follows.
Section~\ref{sec:Preliminaries} provides preliminaries.
Section~\ref{sec:Innate} reviews the computability-theoretic limitation of the LM.
Section~\ref{sec:Negligible} formally states that we can make hallucinations statistically negligible.
Section~\ref{sec:Paradox} solves the paradoxical conflict between the statements provided by Section~\ref{sec:Innate} and Section~\ref{sec:Negligible} through information theory's lens, clarifying that our statement in Section~\ref{sec:Negligible} is more relevant to practical perspectives. 
Section~\ref{sec:Conclusion} concludes the paper.

\subsection{Notation}
We use the symbol $:=$ to define the left-hand side by the right-hand side. We denote the set of real numbers, the set of integers, and the set of nonnegative integers by $\Real$, $\Integer$, and $\Integer_{\ge 0}$, respectively.
We denote the floor function and ceiling function by $\lfloor \cdot \rfloor$ and $\lceil \cdot \rceil$, respectively, i.e., for $a \in \Real$, $\lfloor a \rfloor := \max \{a' \in \Integer \mid a' \le a\}$ and $\lceil a \rceil := \min \{a' \in \Integer \mid a' \ge a\}$.
For a nonnegative integer $\StringLen$ and a set $A$, we denote the direct product set of $\StringLen$ copies of $A$ by $A^{\StringLen}$ (e.g., $A^{3} = A \times A \times A$).
For a set $A$, $2^{A}$ denotes the power set of $A$, i.e., the set of all subsets of $A$. 
Also, $|A|$ denotes the cardinality of $A$.
In particular, the cardinality $|A|$ equals the number of elements in the set $A$ if $A$ is a finite set.

\section{Preliminaries}
\label{sec:Preliminaries}

\begin{definition}[String and the set of strings]
Let $\SymbolSet$ be the set of input symbols.
For example, $\SymbolSet = \{\LetterInstance{A}, \LetterInstance{B}, \dots, \LetterInstance{Z}, \LetterInstance{a}, \LetterInstance{b}, \dots, \LetterInstance{z}, \LetterInstance{.}, \LetterInstance{,}, \LetterInstance{!}, \LetterInstance{?}, \LetterInstance{ }\}$ in the typical English language setting.
A finite-length sequence of symbols is called a \NewTerm{string}.
For $\StringLen \in \Integer_{\ge 0}$, we denote by $\SymbolSet^{\StringLen}$ the direct product set of the $\StringLen$ copies of $\SymbolSet$, i.e., the set of strings of length $\StringLen$.
We denote by $\SymbolSet^{*}$ the set of strings, i.e., $\SymbolSet^{*} := \SymbolSet^{0} \cup \SymbolSet^{1} \cup \dots$. We denote the set of all probability measures on $\SymbolSet^*$ by $\MeasureSetOn (\SymbolSet^*)$.
\end{definition}

For example, \StringInstance{language} $\in \SymbolSet^{8}$ as the word consists of 8 alphabet letters. Likewise, \StringInstance{language model} $\in \SymbolSet^{14}$ as the phrase consists of 14 letters including a space letter.

Below, we define a language model, which is our main focus in this paper.
Since our ultimate motivation is to compare our results with those in \citep{xu2024hallucination}, we adopt the discrete set framework in the chatbot context as they did.
The framework is also compatible with the nature of natural language processing, where discrete natural language tokens are processed.

\begin{definition}[Language model (LM)]
A (deterministic) computable map $\LM: \SymbolSet^{*} \rightarrow \SymbolSet^{*}$ is called a \NewTerm{language model (LM)}.
Here, we say a map $\LM$ is \NewTerm{computable} if there exists a Turing machine halts with just $\LM (\String)$ on its tape for every input $\String$.
We denote the set of all LMs by $\LMSet$. Specifically, $\LMSet := \{\LM: \SymbolSet^{*} \rightarrow \SymbolSet^{*} \mid \text{$\LM$ is computable}\}$.
\end{definition}

Refer to, e.g. \citep{sipser2012introduction}, for rigorous definitions of, e.g., Turing machines.
\begin{remark}[All LLMs are LMs.]
The definition of the computability of a function is invariable even if we replace the Turing machine in the definition with another well-known computation model such as the $\lambda$-calculus, $\mu$-recursive function, or a modern computer with unlimited amounts of time and storage space.
No matter what computing device, training and inference algorithms, and datasets for pre-training and fine-tuning datasets we use, the resulting LM $\LM$ is computable thus in the set $\LMSet$ as long as it is deterministic.
In particular, every LLM (*large* LM) is also in $\LMSet$. 
The computability of LMs plays a crucial role in the discussion in the following section. 
In contrast, since the largeness of a LM itself does not directly matter in this paper, we only use the term LM, not LLM, in the remainder of this paper. Nevertheless, all the discussions concerning LMs in this paper apply to any LLM.
\end{remark}

Take examples to be familiar with notation. If we input \StringInstance{What is language?} $\in \SymbolSet^{17} \subset \SymbolSet^{*}$ to a LM $\LM_{1}$, then it may output \StringInstance{A system of communication.} $\in \SymbolSet^{26} \subset \SymbolSet^{*}$.
In this case, $\LM_{1} (\StringInstance{What is language?}) = \StringInstance{A system of communication.}$
Note that we consider a deterministic map as a LM, so the output of this LM $\LM_{1}$ with the input \StringInstance{What is language?} is always \StringInstance{A system of communication.} and the LM $\LM_{1}$ has no stochastic behavior.

\begin{remark}[Reason for considering deterministic LMs only]
LMs are often defined as a conditional probability mass function $\PMF (\Symbola_{\Time}|\Symbola_{1}, \Symbola_{2}, \dots, \Symbola_{\Time - 1})$, which eventually defines the conditional probability mass function of the output string defined on the output space $\SymbolSet^{*}$.
This definition sees a LM as a stochastic algorithm.
Nevertheless, for simplicity, we focus on deterministic LMs only, which are special cases of stochastic LMs. The reasons why this simple discussion suffices in this paper are the following:
\begin{itemize}
    \item To show the existence of a LM satisfying desirable conditions, which is the main goal of this paper, it suffices to raise a special case.
    \item If we aim to avoid hallucinations, it is a reasonable strategy to make the best output that is known not to be a hallucination, rather than having a possibility of output multiple strings against the given input.
    \item In practice, even when we define a LM as a stochastic algorithm, it often works in practice as a deterministic output algorithm through, e.g., the beam-search algorithm.
    \item It is in line with the setting in \citep{xu2024hallucination}, so it allows us to focus on the essential difference between the previous work and our results.
\end{itemize}
It is, still, interesting to consider the compatibility of reducing hallucinations and output diversity, in which case considering stochastic LMs is beneficial. However, we leave such discussion to future work.
\end{remark}

\begin{remark}[$\SymbolSet^{*}$ and $\LMSet$ are countable.]
Both $\SymbolSet^{*}$ and $\LMSet$ are \NewTerm{countably infinite} sets.
Here, we say an infinite set $A$ is countably infinite if there is a injective map from $A$ to the set of nonnegative integers $\Integer_{\ge 0}$.
The set $\SymbolSet^{*} = \SymbolSet^{0} \cup \SymbolSet^{1} \cup \dots$ of finite-length strings is countably infinite since it is the countably infinite union of finite sets.
The set $\LMSet$ of computable maps are also countably infinite since there exists a universal Turing machine, which emulates any Turing machine from a string describing the machine, so $\LMSet$ can be identified as an infinite subset of $\SymbolSet^{*}$.
The countablity of $\SymbolSet^{*}$ and $\LMSet$ play a core role in Section~\ref{sec:Innate}, in particular in the proof of Theorem~\ref{thm:InfiniteHallucination}.
\end{remark}

We formally define hallucinations. Our definition of hallucinations is not semantic but rather formal so that it is in line with the previous work \citep{xu2024hallucination}. We begin by defining an acceptable output set for each possible input and define hallucinations for each input as the complement set of the acceptable output set.

\begin{definition}[Acceptable outputs and hallucinations]
An \NewTerm{acceptable output set map} is a map $\AcceptableMap: \SymbolSet^{*} \rightarrow 2^{\SymbolSet^{*}}$, i.e., a map taking a string as an input returning a set of strings. 
When we fix an acceptable output set map $\AcceptableMap$, for a string $\String \in \SymbolSet^{*}$ we call $\AcceptableMap (\String)$ the \NewTerm{acceptable output set} for the input string $\String$.
We say that the acceptable output set map $\AcceptableMap$ is \NewTerm{non-vacuous} if $\AcceptableMap (\String) \neq \emptyset$ for all $\String \in \SymbolSet^*$.
We can regard $\AcceptableMap$ as a formulation of the \textbf{ground truth}, and we say that an LM $\LM \in \LMSet$ \NewTerm{hallucinates} on the input $\String \in \SymbolSet^{*}$ with respect to $\AcceptableMap$ if $\LM (\String) \notin \AcceptableMap(\String)$.
\end{definition}

\begin{remark}
When ignore the change of acceptable outputs depending on the era, we can \textbf{fix} an acceptable output set map $\AcceptableMap$, but we will never \textbf{know} the map completely.
Hence, our theoretical interest is in worst-case analysis with respect to the $\AcceptableMap$.
Obviously, it is trivial that we cannot avoid a hallucination for the input $\String$ if $\AcceptableMap (\String) = \emptyset$, so we omit those cases from the consideration.
For some input instances, it has been proved that we cannot directly answer them.
For example, it is known to be impossible to answer by yes/no the question \StringInstance{Is the continuum hypothesis true?} under a widely used axiomatic system (e.g., ZFC) of set theory.
Even in that case, we would say \StringInstance{It can be neither proved nor disproved under ZFC.} $\in \AcceptableMap (\StringInstance{Is the continuum hypothesis true?})$.
Hence, we can assume the existence of $\AcceptableMap$ such that its return value is always nonempty, regardless of G\"odel's incompleteness theorems or the existence of undecidable problems in computability theory.
\end{remark}

\section{Innate computability limitation of LMs}
\label{sec:Innate}

We first formally state the innate limitation of the LMs in the computability aspect.

The following theorem is a modified version of Theorems 2 and 3 in \cite{xu2024hallucination}.

\begin{theorem}
\label{thm:InfiniteHallucination}
There exists an acceptable map $\AcceptableMap: \SymbolSet^{*} \rightarrow 2^{\SymbolSet^{*}}$ such that
\begin{itemize}
    \item $|\AcceptableMap (\String)| > 0$ for every $\String \in \SymbolSet^{*}$, and
    \item For any $\LM \in \LMSet$, $\LM$ hallucinates on infinitely many inputs, i.e., $\{\String \in \SymbolSet^{*} \mid \LM(\String) \notin \AcceptableMap (\String)\}$ is an infinite set.
\end{itemize}
\end{theorem}

\begin{proof}
See Appendix~\ref{sec:ProofHallucination}.
\end{proof}

Theorem~\ref{thm:InfiniteHallucination} claims that in the worst case with respect to the acceptable map $\AcceptableMap$, no matter what LM we use, it hallucinates on infinitely many input strings.
Note that this negative result holds regardless of our choice of neural network architecture, algorithms, and training data.

\begin{remark}
Theorem~\ref{thm:InfiniteHallucination} is similar to
Theorems 2 and 3 in \cite{xu2024hallucination} both in its statement and its proof strategy but technically stronger than those for the following senses: 
\begin{itemize}
    \item Our Theorem~\ref{thm:InfiniteHallucination} claims the existence of a map $\AcceptableMap$ for which all the LMs hallucinates on infinitely many input sequences. In particular, $\AcceptableMap$ does not depend on the choice of $\LM$.
    Theorem 2 in \cite{xu2024hallucination} does not consider the whole LMs in $\LMSet$ and Theorem 3 in \cite{xu2024hallucination} allows the dependency of $\AcceptableMap$ on the choice of the learning procedure.
    Nevertheless, if we note that $\LMSet$ is a countable set, the modification from Theorems 2 and 3 in \cite{xu2024hallucination} to our Theorem~\ref{thm:InfiniteHallucination} is straightforward.
    \item Our proof avoids using the axiom of choice. See Appendix~\ref{sec:ProofHallucination} for details, including why it matters in computer science, not in the context of pure mathematics. 
\end{itemize}
\end{remark}

Theorem~\ref{thm:InfiniteHallucination} may look fatally negative to practitioners at one glance as it states that infinite hallucinations are inevitable. However, our main claim is that this result itself is not a practical issue, as we explain below.

\section{Hallucinations can be made statistically negligible}
\label{sec:Negligible}

As a preliminary, we first formalize a training data sequence and a LM trainer, which receives a training data sequence and returns a LM.

\begin{definition}[Training dataset and language model trainer]
\label{def:LMT}
An input-output string pair $(\String, \Output) \in \SymbolSet^{*} \times \SymbolSet^{*}$ is called a \NewTerm{training data point}. Also, a finite sequence \\ $((\String_{1}, \Output_{1}), (\String_{2}, \Output_{2}), \dots, (\String_{\DataLen}, \Output_{\DataLen})) \in (\SymbolSet^{*} \times \SymbolSet^{*})^{*}$ of training data points is called a \NewTerm{training data sequence} or \NewTerm{training dataset}.
A map $\Trainer: (\SymbolSet^{*} \times \SymbolSet^{*})^{*} \to (\SymbolSet^{*} \to \SymbolSet^{*})$, taking a training data sequence as and input and returning a LM, is called a \NewTerm{language model trainer (LMT)}.
\end{definition}

\begin{remark}
Any practical LM can be regarded as an output of a computable LMT. This includes cases where a neural network is pretrained first on a general corpus to make a general next token predictor and fine-tuned on input-output string pairs (a training dataset) to modify the neural network model to one for a chat-bot. In practice, it is sufficient to consider \textbf{computable} LMTs. Nevertheless, we do not assume the computability of a LMT to clarify that the computability does not essentially matter in the following statistical results. Obviously, the theoretical results holding on general LMTs also apply to computable LMTs. 
\end{remark}

In this paper, we are interested in the probability of hallucinations happening, rather than the number of input instances causing hallucinations.
Hence, we formally define the hallucination probability.

\begin{definition}[Hallucination probability]
We define the \NewTerm{hallucination probability} $\HallucinationProbability_{\Measure} (\LM) \in [0, 1]$ of a LM $\LM$ on a probability measure $\Measure$ on $\SymbolSet^{*}$ by
\[
\HallucinationProbability_{\Measure} (\LM) := \Probability (\LM (\StringRV) \notin \AcceptableMap(\StringRV)),
\]
where the right-hand side is the probability with respect to the random variable $\StringRV$ generated by $\Measure$.  
\end{definition}

\begin{remark}
The hallucination probability is often called the \NewTerm{0-1 risk} in a general statistical learning theory context.
\end{remark}

Now, let's clarify our goal.
The main claim of the inevitability results by \citep{xu2024hallucination} and \citep{banerjee2024llms} is that hallucinations are inevitable regardless of the choice of training and inference algorithms and training data sequence (dataset) if we consider the worst case with respect to the ground truth acceptable output set map $\AcceptableMap$.
To argue that they do not have practical implications from statistical aspects, we need to prove that there exists an algorithm and a desired property of datasets such that for any acceptable output set map $\AcceptableMap$ and probability measure $\Measure$, we can make the hallucination probability arbitrarily small.
Here, to consider the logical negation, we can consider any algorithm and property of datasets, but we should not assume anything about the acceptable output set map $\AcceptableMap$ or the probability measure $\Measure$. Of course, we can always obtain a good consequence if we make a strong assumption, but such a consequence cannot disprove the inevitability results by \citep{xu2024hallucination} and \citep{banerjee2024llms} unless we can confirm that real natural language satisfies it.
Practically, there is no mathematically tractable assumption that real natural language satisfies. 

\begin{example}[We cannot assume that the ground truth is computable]
For example, if we know that there is some countable set, e.g., the set of computable maps, that the acceptable output set map $\AcceptableMap$ belongs to, learning is possible from Mark Gold's classical discussion \citep{gold1967language}.
However, we have no guarantee, for example, that $\AcceptableMap$ is computable.
Indeed, the core of the diagonal arguments \citep{xu2024hallucination,banerjee2024llms} is the uncountability of the set of hypotheses (in this paper's notation, the set that $\AcceptableMap$ may belong to).
Saying that "if we limit the search space to some countable set, then learning is possible" is not a valid rebuttal to the claim that "since the search space is uncountable, hallucination is inevitable."
\end{example}

For the above reason, we are going to find an ideal property of a training data sequence, denoted by $\TrainingDataRV$ in this paper and an appropriate training and inference algorithm, denoted by $\Trainer$, exploiting the training data sequence.
We begin by defining an ideal property of a training data sequence.

\begin{definition}[Qualified random training data sequence]
\label{def:qualified}
Assume that $\AcceptableMap (\String) \neq \emptyset$ for all $\String \in \SymbolSet^{*}$ and let $\Measure$ be a probability measure on $\SymbolSet$.
Then, a $(\SymbolSet^{*} \times \SymbolSet^{*})^{\DataLen}$-valued random variable $\TrainingDataRV = ((\StringRV_{1}, \OutputRV_{1}), (\StringRV_{2}, \OutputRV_{2}), \dots, (\StringRV_{\DataLen}, \OutputRV_{\DataLen}))$ is called a length-$\DataLen$ \NewTerm{qualified random training data sequence} compatible with $\AcceptableMap$ generated by $\Measure$ if $\TrainingDataRV$ is generated as follows:
\begin{itemize}
    \item $\StringRV_{1}, \StringRV_{2}, \dots, \StringRV_{\DataLen}$ are $\SymbolSet^{*}$-valued random variables independently and identically generated by $\Measure$.
    \item For $\IDatum = 1, 2, \dots, \DataLen$, the distribution of the $\SymbolSet^{*}$-valued random variable $\OutputRV_{\IDatum}$ is determined only by the value of $\StringRV_{\IDatum}$ and $\OutputRV_{\IDatum} \in \AcceptableMap (\StringRV_{\IDatum})$ is satisfied in probability 1.
\end{itemize}
\end{definition}

\begin{remark}[Qualified random training data sequence assumption's strength]
\
\\
From practical aspects, it is a strong assumption to presume $\OutputRV_{\IDatum} \in \AcceptableMap (\StringRV_{\IDatum})$ is satisfied in probability 1. In fact, the following discussion essentially holds as long as the most frequently appearing output string is in the acceptable set. 
However, this extension bring to discussion complexity unnecessary for our motivation, which is to clarify the computability-based limitation of LMs is not a practical issue. Hence, we omit such an extension.
\end{remark}

Then, we formally define statistical negligibility.
Our definition is inspired by the framework of probably approximately correct (PAC) learning, but our viewpoint is from hallucinations and distributions, rather than from the hypothesis set. 
This is to make easier its comparison to the result in Section~\ref{sec:Innate}.

\begin{definition}[Statistical negligiblity of hallucinations]
\label{def:negligible}
(1) We say that hallucinations of a LMT $\Trainer$ with a qualified random training data sequence are $(\epsilon_{\HallucinationSymbol}, \epsilon_{\TrainSymbol})$-\NewTerm{negligible} on $\Measure \in \MeasureSetOn(\SymbolSet^*)$ with training sequence length $\bar{m}$ if for any non-vacuous acceptable output set map $\AcceptableMap$, any $\DataLen \ge \bar{\DataLen}$, and any length-$\DataLen$ qualified random training data sequence $\TrainingDataRV$ compatible with $\AcceptableMap$ generated by $\Measure$, the hallucination probability satisfies $\HallucinationProbability_{\Measure} (\Trainer (\TrainingDataRV)) < \epsilon_{\HallucinationSymbol}$ in probability (with respect to $\TrainingDataRV$) at least $1 - \epsilon_{\TrainSymbol}$.

(2) Let $\MeasureSet \subset \MeasureSetOn (\SymbolSet^*)$ be a set of probability measures on $\SymbolSet^*$. 
For a set of probability measure $\MeasureSet \subset \MeasureSetOn (\SymbolSet^*)$, we say that hallucinations of a LMT $\Trainer$ with a qualified random training data sequence are \NewTerm{uniformly statistically negligible} on $\MeasureSet$ if for any $\epsilon_{\HallucinationSymbol}, \epsilon_{\TrainSymbol} \in (0, 1]$ there exists a $\bar{\DataLen} \in \Integer_{\ge 0}$ such that for any $\Measure \in \MeasureSet$, hallucinations are $(\epsilon_{\HallucinationSymbol}, \epsilon_{\TrainSymbol})$-negligible on $\Measure$ with training sequence length $\bar{m}$.
Also, we say that hallucinations of a LMT $\Trainer$ with a qualified random training data sequence are \NewTerm{non-uniformly statistically negligible} on $\MeasureSet$ if for any $\epsilon_{\HallucinationSymbol}, \epsilon_{\TrainSymbol} \in (0, 1]$ and any $\Measure \in \MeasureSet$, there exists a $\bar{\DataLen} \in \Integer_{\ge 0}$ such that hallucinations are $(\epsilon_{\HallucinationSymbol}, \epsilon_{\TrainSymbol})$-negligible on $\Measure$ with training sequence length $\bar{m}$.
\end{definition}

\begin{remark}[Computability conditions and statistical negligibility]
It is known \\ \citep{agarwal2020learnability, agarwal2021open, sterkenburg2022characterizations} that whether or not a function class is PAC‑learnable can depend on whether the learning algorithm is required to be computable. Similarly, when we ask if hallucinations can be made statistically negligible over a distribution class, the answer may also hinge on assuming the algorithm is computable. 
However, in our paper, the positive results showing statistical negligibility are always obtained by explicitly constructing a computable algorithm, and the negative results, demonstrating failure to achieve ineligibility, are based on the no‑free‑lunch theorem that remains valid even when considering potentially non‑computable procedures. 
As a result, our theoretical conclusions about statistical negligibility hold, regardless of whether the algorithms are required to be computable.
\end{remark}

\begin{remark}[Meaning of statistical negligibility and its uniformity]
If \\ hallucinations are statistically negligible, it implies that we can make the probability of hallucinations arbitrarily small with the help of a qualified and sufficiently long training sequence.
It is because we can choose arbitrarily small $\epsilon_{\HallucinationSymbol}$, and $\epsilon_{\TrainSymbol}$.
The difference between the above uniform statistical negligibility and non-uniform statistical negligibility of hallucinations only lies in whether the training data length $\bar{\DataLen}$ can depend on the probability measure $\Measure$ or not.
Specifically, if hallucinations are \Emph{uniformly} statistically negligible, we know in advance a sufficient condition on the training data size $\DataLen$. 
In contrast, if we only know hallucinations are \Emph{non-uniformly} statistically negligible, we do not know how long a training data sequence we need, but eventually we can achieve the aimed hallucination probability (with high probability over training data distribution) if we increase the data size.
By definition, if hallucinations are uniformly statistically negligible, then non-uniformly statistically negligible.
\end{remark}

Interestingly, as shown in the Appendix, no LMT $\Trainer$ can achieve the uniform statistical negligibility on the set $\MeasureSetOn(\SymbolSet^*)$ of all probability measures. In contrast, we can achieve non-uniform statistical negligibility. 
While we will state positive results for both later, as a preliminary for results about the uniform statistical negligibility, we define below the cumulative distribution function of the input length.

\begin{definition}[Cumulative distribution function (CDF) of the input length]
For a probability measure $\Measure$ on $\SymbolSet^{*}$, we denote by $\CDF_{\Length \sharp \Measure}$ the cumulative distribution function (CDF) of the length of a random variable generated by $\Measure$.
Specifically, $\CDF_{\Length \sharp \Measure}: \Integer_{\ge 0} \rightarrow [0, 1]$ is defined by $\CDF_{\Length \sharp \Measure} (\StringLen) := \Probability (\Length (\StringRV) \le \StringLen)$, where $\StringRV$ is generated by $\Measure$.
\end{definition}

\begin{definition}[The set of probability measures with a CDF lower bound]
\label{def:BoundedMeasures}
Fix a non-decreasing function $\overline{\CDF}: \Integer_{\ge 0} \rightarrow [0, 1]$ that satisfies $\lim_{\StringLen \to +\infty} \overline{\CDF} (\StringLen) = 1$. 
We denote by $\MeasureSet_{\overline{\CDF}}$ the set of probability measures whose input length CDF is lower-bounded by $\overline{\CDF}$, defined as
\[
\MeasureSet_{\overline{\CDF}}
:= \{\Measure \mid \forall n \in \Integer_{\ge 0}, \CDF_{\Length \sharp \Measure} (\StringLen) \ge \overline{\CDF} (\StringLen)\}.
\]  
\end{definition}

Now, we are ready to state our main result.
\begin{theorem}[Hallucinations can be statistically negligible with $\overline{\CDF}$]
\label{thm:negligible}
There exists a LMT $\Trainer$ (defined in Definition~\ref{def:LMT}) such that:
\begin{enumerate}
    \item for any non-decreasing function $\overline{\CDF}: \Integer_{\ge 0} \rightarrow [0, 1]$ that satisfies $\lim_{\StringLen \to +\infty} \overline{\CDF} (\StringLen) = 1$, hallucinations of $\Trainer$ with a qualified random training data sequence are \Emph{uniformly statistically negligible} on $\MeasureSet_{\overline{\CDF}}$ in the sense of Definition~\ref{def:negligible} and
    \item hallucinations of $\Trainer$ with a qualified random training data sequence are \Emph{non-uniformly statistically negligible} on the set $\MeasureSetOn (\SymbolSet^*)$ of all the probability measures on $\SymbolSet^*$ in the sense of Definition~\ref{def:negligible}.
\end{enumerate}
\end{theorem}

\begin{proof}
Since for any $\Measure \in \MeasureSetOn (\SymbolSet^*)$, $\Measure \in \MeasureSet_{\CDF_{\Length \sharp \Measure}}$, we have that $\MeasureSetOn (\SymbolSet^*) = \bigcup_{\Measure \in \MeasureSetOn (\SymbolSet^*)} \MeasureSet_{\CDF_{\Length \sharp \Measure}}.$
Hence, (2) follows from (1).
(1) follows from Proposition~\ref{prp:trivial} immediately, which is stated in Appendix.
\end{proof}

\begin{remark}[Summary of the assumptions for Theorem~\ref{thm:negligible}]
For benefits of readers, we summarize the assumptions of Theorem~\ref{thm:negligible}.
Theorem~\ref{thm:negligible} claims that hallucinations are non-uniformly statistically negligible (regardless of $\AcceptableMap$ and $\Measure$) if a certain LMT is used, when all of the following conditions are met.
\begin{itemize}
    \item Training data is qualified in the sense of Definition~\ref{def:qualified}.
    \item Training data is sufficient in the sense that it is more than $\bar{\DataLen}$ in Definition~\ref{def:negligible}.
\end{itemize}
Also, hallucinations are uniformly statistically negligible if the following condition is also met.
\begin{itemize}
    \item We know some lower bound of the input length CDF.
\end{itemize}
We remark that this does not require us to know any information about natural language's grammar or syntax.
\end{remark}

\begin{remark}
\label{rem:HugeDataSize}
Although specific data size $\bar{\DataLen}$ is not mentioned in the statement of Theorem~\ref{thm:negligible}, $\bar{\DataLen}$ can be huge, and such a huge training data size is inevitable under this paper's framework, as we will discuss in Appendix~\ref{sec:DataSize}. However, this fact is not meant to assert that an extraordinary amount of data is practically necessary for the success of LMs. Rather, the framework of this paper is intended to rebut the practical implication of the "inevitability of hallucination" based on a diagonal argument. Its purpose is to conclude that even when assumptions are stripped down to this extent to align with their framework (for instance, even without assuming the computability of the acceptable output set map $\AcceptableMap$), hallucinations can still be made statistically negligible. 
Naturally, if certain practical assumptions can be made, the sufficient data length can be shorter.
However, we avoid such assumptions in this paper to clarify the essence of our rebuttal to diagonal-argument-based negative results. 
\end{remark}

\Emph{Further discussions in the Appendix}: Although the form of Theorem~\ref{thm:negligible} suffices to rebut the negative results in\citep{xu2024hallucination}, in the Appendix, we provide the specific form of Theorem~\ref{thm:negligible} and indicate that the sufficient training data size can be huge, as stated in Remark~\ref{rem:HugeDataSize}. 
We also show that the specific theorem is nearly optimal in that its assumptions regarding the training data size and the availability of an input length CDF lower bound cannot be removed. 
We also point out that the optimality of our theorem enlightens future work directions.

Now, we have provided Theorem~\ref{thm:negligible} stating that the probability of hallucinations can be arbitrarily small, which is a positive result from the probability theory aspect.
However, the negative result Theorem~\ref{thm:InfiniteHallucination}, stating that hallucinations happen on infinite input instances still holds even in this case.
The above positive and negative results seem to contradict each other.
How do we interpret these seemingly contradicting two results from the viewpoint of practice?
The following section answers this question.

\section{Paradox and solution: infinite input instances causing hallucinations but with arbitrarily small probability}
\label{sec:Paradox}

This section, first, clarifies why the paradoxical conflict between the statements provided by Section~\ref{sec:Innate} and Section~\ref{sec:Negligible} coexist, and argue that our statement in Section~\ref{sec:Negligible} is more relevant to practical perspectives through information theory's lens.

\subsection{Why can those seemingly contradicting results coexist?}
One might feel the negative result of Theorem~\ref{thm:InfiniteHallucination} and the positive statement of Theorem~\ref{thm:negligible} contradict each other.
Indeed, since Theorem~\ref{thm:InfiniteHallucination} makes no assumption on the data distribution, it still applies to the setting of Theorem~\ref{thm:negligible}. 
Hence, under the same setting, Theorem~\ref{thm:InfiniteHallucination} states that every LM hallucinates on infinite input instances, whereas Theorem~\ref{thm:negligible} states that there is a LM, of which hallucinations are statistically negligible.
In fact, they do not contradict each other mathematically.
Since the support $\SymbolSet^{*}$ of the probability measures that we consider is an infinite set, the infinite subset on which a LM hallucinates can have little probability.
An intuitive example of an infinite set having an arbitrarily small probability is the set $\Integer_{\ge m} = \{m, m+1, \dots\}$ that has probability $(1/2)^m$ when nonnegative integer $i$ has probability mass $(1/2)^i$. Here, $\Integer_{\ge m}$ is an infinite set for any fixed $m$, but its probability converges to 0 as $m$ increases.
Likewise, for fixed $\DataLen$, the set of input instances on which a LM hallucinates is an infinity set as stated in Theorem~\ref{thm:InfiniteHallucination}, but we can make the \Emph{probability} of the set converge to zero as $\DataLen$ increases, as suggested by Theorem~\ref{thm:negligible}.

\subsection{Infinite set, but with arbitrarily small probability. Which matters in practice?}
Now, what should be discussed is whether the infinite but arbitrarily small probability errors are accepted in practice.
This is no longer a mathematical discussion and can ultimately depend on the domain.
Nevertheless, we still claim that it has practically been negligible in information theory, where Shannon's source coding theorem is one of its foundations.
Here, one of the most fundamental versions of Shannon's source coding theorem states the following (appearing in, e.g., \cite{mackay2003information}).

\begin{theorem}[Shannon's source coding theorem]
\label{thm:source-coding}
Consider a probability measure $\Measure$ on a finite set $\mathcal{X}$ and suppose that its entropy is $H$ bits. Denote the product set of the $\DataLen$ copies of $\mathcal{X}$ by $\mathcal{X}^{\DataLen}$ and the product measure of $\DataLen$ copies of $\Measure$ by $\Measure^{\DataLen}$. In other words, $\Measure^{\DataLen}$ is the probability measure generating a random variable sequence $X_{1} X_{2} \dots X_{\DataLen}$, where $X_{i}$ is generated by $\Measure$ for $i = 1, 2, \dots, \DataLen$. Given $\epsilon > 0$ and $0 < \delta < 1$, there exists a positive integer $\DataLen_0$ such that for any positive integer $\DataLen > \DataLen_0$,
there exists a set $A_{\DataLen} \subset \mathcal{X}^{\DataLen}$ such that 
\begin{itemize}
    \item $|\frac{1}{\DataLen} \log_{2} |A_{\DataLen}| - H| < \epsilon$, and
    \item $\Measure^{\DataLen} (A_{\DataLen}) > 1 - \delta$.
\end{itemize}
\end{theorem}

Here, the first bullet point in Theorem~\ref{thm:source-coding} indicates that the number of elements in $A_{\DataLen}$ is smaller than $2^{\DataLen (H + \epsilon)}$ and so $\lceil\DataLen (H + \epsilon)\rceil$ bits are sufficient to code every element in $A_{\DataLen}$. The second bullet point indicates that the elements not in $A_{\DataLen}$ appear in probability at most $\delta$. Here, note that the number of elements of the set $A_{\DataLen}$ is much smaller than that of $\mathcal{X}^{\DataLen}$.
As we can set $\epsilon$ and $\delta$ arbitrarily, the theorem has been understood as follows (the statement appears in the first half of ``verbal statement'' Shannon's source coding theorem in \cite{mackay2003information}).

\begin{theorem}[Shannon's source coding theorem (verbal statement)]
$\DataLen$ i.i.d. random variables each with entropy $H$ can be compressed into more than $m H$ bits with negligible risk of information loss, as $\DataLen \to \infty$.
\end{theorem}

The above common understanding of Shannon's source coding theorem implies that, if we can make the probability of some unpreferred event arbitrarily small, the event is considered to be practically negligible in information theory even if the number of elements in the event is large.

Therefore, we can conclude that, \textbf{although infinite hallucinations are inevitable in the sense of Theorem~\ref{thm:InfiniteHallucination}, they can be practically negligible in the application domains where information theory does not cause a practical issue.}

\section{Conclusion}
\label{sec:Conclusion}

We have shown that hallucinations are statistically negligible with an appropriate algorithm if the quality and quantity of the training data are sufficient, even in the worst scenario with respect to the ground truth and distribution.
While hallucinations on an infinite set of inputs cannot be entirely eliminated, their probability can always be reduced by improving algorithms and training data. Section~\ref{sec:Paradox} has also pointed out that the hallucination probability, rather than the "number" of hallucinations, reflects practical considerations. 
By synthesising the above discussions, we can say that there exists a combination of dataset and algorithm that can make hallucinations unproblematic in practice. 
The existence of such a combination of dataset and algorithm (regardless of whether it is realistically preparable) implies the following: \Emph{if hallucinations are indeed a practical issue, the cause should be attributed to either the dataset (quality or quantity) or the algorithm (including issues of computational complexity), and not to an "innate" inevitability of hallucinations derived from computability theory based on the diagonal argument}.

\acks{Yulan He was supported by the UK Engineering and Physical Sciences Research Council (EPSRC) through a Turing AI Fellowship (grant no. EP/V020579/1, EP/V020579/2). Zhongyuan Wang is supported by  National Natural Science Foundation of China 62371350.
}

\newpage
\appendix
\section{Proof and optimality of Theorem~\ref{thm:negligible}}
\label{sec:ProofPositive}

In this section, we discuss the assumptions of our main theorem Theorem~\ref{thm:negligible}.
Theorem~\ref{thm:negligible} assumes that the input length CDF lower bound is available and also that the training data size, as we see in detail later.
We first go through our proof strategy based on constructing a trivial algorithm to see intuitively where those assumptions come from.
Then, we theoretically show that those assumptions are necessary.
The implication is significant since it means that no matter what algorithms we use, the worst-case data size is almost the same as that with which a trivial algorithm can succeed.
We conclude this section by remarking that our result implies that we must not try to universally succeed under loose assumptions and rather should make stronger assumptions reflecting the nature of natural languages.

\subsection{Proof strategy for Definition~\ref{def:negligible}}
We construct a trivial algorithm Algorithm~\ref{alg:trivial}, named Rote Memorizer (RM), to prove Theorem~\ref{thm:negligible}, stating that hallucinations can be statistically negligible. 
The algorithm gives us intuition behind the assumptions of Theorem~\ref{thm:negligible}.
The idea of the algorithm is simple. We first find an input length threshold $\bar{\StringLen}$ only depending on $\DataLen$ and $\overline{\CDF}$, and we simply rote-memorize the input-output pairs with an input shorter than $\bar{\StringLen}$ in the training data sequence, 
If the training data size $\DataLen$ is sufficient, there exists a string length $\bar{\StringLen}$ such that it is so short that all the input strings shorter than $\bar{\StringLen}$ appear in high probability in training data and so long that the probability of the input length longer than $\bar{\StringLen}$ is small.
The pseudocode of the straightforward algorithm is given in Algorithm~\ref{alg:trivial}.
Note that we do NOT insist that Algorithm~\ref{alg:trivial} should be used in practice. It is rather a tool for the proof.

\begin{algorithm}[H]
    \caption{RoteMemorizer (RM)}
    \label{alg:trivial}
    \begin{algorithmic}[1]
        \Procedure{RoteMemorizer}{Training data $((s_{1}, y_{1}), (s_{2}, y_{2}), \dots, (s_{\DataLen}, y_{\DataLen}))$}
            \State $\DataLen \gets \Length ((s_{1}, y_{1}), (s_{2}, y_{2}), \dots, (s_{\DataLen}, y_{\DataLen}))$
            \State Initialize an empty dictionary $d$
            \For{$i \gets 1, 2, \dots, \DataLen$}
                \State $d[s_{i}] \gets y_{i}$
            \EndFor
            \Function{TrivialRecaller}{$s \in \SymbolSet^*$}
                \If{$s$ is a key in $d$}
                    \State \textbf{return} $d[s] \in \SymbolSet^*$
                \Else
                    \State \textbf{return} \StringInstance{} $\in \SymbolSet^*$
                \EndIf
            \EndFunction
            \State \textbf{return} the function \textsc{TrivialRecaller}
    \EndProcedure
    \end{algorithmic}
\end{algorithm}

With the help of Algorithm~\ref{alg:trivial}, we can show the following, which immediately gives us Theorem~\ref{thm:negligible}. 

\begin{proposition}
\label{prp:trivial}
Assume that $\AcceptableMap (\String) \neq \emptyset$ for all $\String \in \SymbolSet^{*}$.
Fix a non-decreasing function $\overline{\CDF}: \Integer_{\ge 0} \rightarrow [0, 1]$ that satisfies $\lim_{\StringLen \to +\infty} \overline{\CDF} (\StringLen) = 1$. 
Then, hallucinations of $\Trainer$ given by Algorithm~\ref{alg:trivial} with qualified random training data sequence are uniformly statistically negligible in the sense of Definition~\ref{def:negligible} on $\MeasureSet_{\overline{\CDF}}$, defined by Definition~\ref{def:BoundedMeasures}.
Here, $\bar{\DataLen}$ in the definition of uniform statistical negligibility is given by 
$\bar{\DataLen} = \lceil\frac{|\SymbolSet^{*}|^{(\bar{\StringLen} + 1)}}{1 - \overline{\CDF} (\bar{\StringLen})} \ln \frac{|\SymbolSet^{*}|^{(\bar{\StringLen} + 1)}}{2(1 - \overline{\CDF} (\bar{\StringLen}))}\rceil$, where $\bar{\StringLen}$ is an integer that satisfies $1 - \overline{\CDF} (\bar{\StringLen}) < \frac{1}{2} \min \{\epsilon_{\HallucinationSymbol}, \epsilon_{\TrainSymbol}\}$.
Note that such a $\bar{\StringLen}$ exists by the definition of $\overline{\CDF}$.
\end{proposition}

\begin{proof}
See Appendix~\ref{sec:ProofTrivial}.
\end{proof}

We can see that for uniform statistical negligibility of hallucinations, Proposition~\ref{prp:trivial} requires 
\begin{itemize}
    \item the input length CDF lower bound and 
    \item the huge size $\bar{\DataLen} = \lceil\frac{|\SymbolSet^{*}|^{(\bar{\StringLen} + 1)}}{1 - \overline{\CDF} (\bar{\StringLen})} \ln \frac{|\SymbolSet^{*}|^{(\bar{\StringLen} + 1)}}{2(1 - \overline{\CDF} (\bar{\StringLen}))}\rceil$ of training data, which is exponential with respect to $\bar{\StringLen}$ depending on $\overline{\CDF}$.
\end{itemize}
From the construction of Algorithm~\ref{alg:trivial}, we can intuitively understand why Proposition~\ref{prp:trivial} requires these two for hallucinations to be uniformly statistically negligible by Algorithm~\ref{alg:trivial}.
To achieve the uniform statistical negligibility, we need to determine the training data size $\DataLen$ in advance.
For the rote memorization algorithm to achieve this, we need to limit the input length. 
With a lower bound of the input length CDF, we can find a string length threshold (which corresponds to $\bar{\StringLen}$) such that we can safely ignore inputs whose length is longer than the threshold, and we can determine the training data size $\DataLen$ so that the training data covers the strings whose length is shorter than the threshold.
Without the lower bound of the input length CDF, we cannot do this.
Also, since it tries to rote memorize all the strings shorter than the string length threshold $\bar{\StringLen}$, it is natural that it requires the training data size exponential to $\bar{\StringLen}$.

A natural question is whether or not we can omit these two assumptions for uniform statistical negligibility of hallucinations from Proposition~\ref{prp:trivial} (or Theorem~\ref{thm:negligible}) by applying possible cleverer algorithms than the trivial Algorithm~\ref{alg:trivial}. 
Interestingly, neither of them can be omitted, as we will see for the input length CDF lower bound in Appendix~\ref{sec:InputLength} and for a huge training data size in Appendix~\ref{sec:DataSize}.

\subsection{The input length CDF lower bound is necessary for uniform statistical negligibility}
\label{sec:InputLength}

The following theorem formally states that for uniform statistical negligibility of hallucinations, we cannot omit the input length CDF lower bound condition, no matter what algorithms we consider. Note that in the following, $\Expect_{\TrainingDataRV}$ and $\Probability_{\TrainingDataRV}$ denote the operators returning the expectation and probability over the length-$\DataLen$ qualified random training data sequence $\TrainingDataRV$.

\begin{theorem}[No free lunch theorem in LM context]
\label{thm:NFLLM}
For any LMT $\Trainer: (\SymbolSet^{*} \times \SymbolSet^{*})^{*} \rightarrow (\SymbolSet^{*} \rightarrow \SymbolSet^{*})$, any training data sequence length $\DataLen \in \Integer_{\ge 0}$, and any $\LowerBound_{\HallucinationSymbol} \in (0, 1)$, there exist a map $\AcceptableMap: \SymbolSet^{*} \rightarrow 2^{\SymbolSet^{*}}$ satisfying $\AcceptableMap (\String) \neq \emptyset$ for all $\String \in \SymbolSet^{*}$ and a probability distribution $\Measure$ on $\SymbolSet^{*}$ such that the hallucination probability satisfies 
\begin{align*}
\Expect_{\TrainingDataRV} \HallucinationProbability_{\Uniform (\underlineDomain), \AcceptableFunc} (\Learner (\TrainingDataRV)) &\ge 1/2, \\
\Probability_{\TrainingDataRV} (\HallucinationProbability_{\Uniform (\underlineDomain), \AcceptableFunc} (\Learner (\TrainingDataRV)) \ge \LowerBound_{\HallucinationSymbol}) &\ge \LowerBound_{\TrainSymbol}:= \frac{1 - 2 \LowerBound_{\HallucinationSymbol}}{2 - 2 \LowerBound_{\HallucinationSymbol}},
\end{align*}
where $\TrainingDataRV$ is the length-$\DataLen$ qualified random training data sequence.
\end{theorem}

See Appendix~\ref{sec:ProofNegative} for the proof of Theorem~\ref{thm:NFLLM}. It uses a variant of the no-free-lunch theorem, which we state in Appendix~\ref{sec:ProofNegative} as Theorem~\ref{thm:NFL} \citep{shalev2014understanding}.

\begin{remark}
For example, if $\LowerBound_{\HallucinationSymbol} = 1/4$, then $\LowerBound_{\TrainSymbol} = 1/3$.
Hence, Theorem~\ref{thm:NFLLM} states that $\HallucinationProbability (\Trainer (\TrainingDataRV)) > 1/4$ happens in probability at least $1/3 - \epsilon$ (for an arbitrarily small $\epsilon$) over the choice of the training data $\TrainingDataRV$ in the worst case on the choice of $\AcceptableMap (\String)$ and $\Measure$.
\end{remark}

Despite its negative statement, we do not consider Theorem~\ref{thm:NFLLM} to be implying issues from a practical perspective, since the lower bound could be easily obtained as it is nothing but a probability distribution of a scalar random variable without any information about syntax or semantics of natural languages.
Nevertheless, Theorem~\ref{thm:NFLLM} is theoretically interesting as proof of an optimality of Theorem~\ref{thm:negligible}.

\subsection{Discussion on training data size}
\label{sec:DataSize}
Proposition~\ref{prp:trivial} suggests that a training data size that is exponential to $\bar{\DataLen}$, which depends on $\overline{\CDF}$, is \Emph{sufficient} to make hallucinations statistically negligible.
The following theorem states its converse in some sense.
Specifically, it says that a training data size that is exponential to $\underline{\DataLen}$, another variable depending on $\overline{\CDF}$, is \Emph{necessary}.

\begin{theorem}[No free lunch theorem in LM context]
\label{thm:NFLLMDataSize}
Fix a non-decreasing function $\overline{\CDF}: \Integer_{\ge 0} \rightarrow [0, 1]$ that satisfies $\lim_{\StringLen \to +\infty} \overline{\CDF} (\StringLen) = 1$,
and define 
\begin{align*}
\underline{\StringLen} &:= \argmin_{\StringLen \in \Integer_{\ge 0}} \frac{|\SymbolSet|^{\StringLen + 1} - 1}{(|\SymbolSet| - 1) \overline{\CDF} (\StringLen)}, \\
\underline{\DataLen} &:= \left\lfloor\frac{|\SymbolSet|^{\underline{\StringLen} + 1} - 1}{(|\SymbolSet| - 1) \overline{\CDF} (\underline{\StringLen})}\right\rfloor.
\end{align*}
For any LMT $\Trainer: (\SymbolSet^{*} \times \SymbolSet^{*})^{*} \rightarrow (\SymbolSet^{*} \rightarrow \SymbolSet^{*})$, any training data sequence length $\DataLen \le \underline{\DataLen}$, and $\LowerBound_{\HallucinationSymbol} \in (0, 1)$, there exist a map $\AcceptableMap: \SymbolSet^{*} \rightarrow 2^{\SymbolSet^{*}}$ satisfying $\AcceptableMap (\String) \neq \emptyset$ for all $\String \in \SymbolSet^{*}$ and a probability distribution $\Measure$ on $\SymbolSet^{*}$ such that the hallucination probability satisfies 
\begin{align*}
\Expect_{\TrainingDataRV} \HallucinationProbability_{\Uniform (\underlineDomain), \AcceptableFunc} (\Learner (\TrainingDataRV)) &\ge 1/2, \\
\Probability_{\TrainingDataRV} (\HallucinationProbability_{\Uniform (\underlineDomain), \AcceptableFunc} (\Learner (\TrainingDataRV)) \ge \LowerBound_{\HallucinationSymbol}) &\ge \LowerBound_{\TrainSymbol}:= \frac{1 - 2 \LowerBound_{\HallucinationSymbol}}{2 - 2 \LowerBound_{\HallucinationSymbol}},
\end{align*}
where $\TrainingDataRV$ is the length-$\DataLen$ qualified random training data sequence.
\end{theorem}

See Appendix~\ref{sec:ProofNegative} for the proof of Theorem~\ref{thm:NFLLMDataSize}. Again, it uses the no-free-lunch theorem Theorem~\ref{thm:NFL} \citep{shalev2014understanding}, which we state in Appendix~\ref{sec:ProofNegative}.

\begin{remark}[Implications of Theorem~\ref{thm:NFLLMDataSize}]
\label{rmk:NFLLMDataSize}
The huge data sizes that appear in Proposition~\ref{prp:trivial} and Theorem~\ref{thm:NFLLMDataSize} do NOT imply that such vast quantities of training data are necessary for language models to succeed in practice.
Rather, it enlightens future work directions, suggesting the necessity of stronger assumptions reflecting the behavior of natural languages.
Specifically, from a theoretical perspective, Theorem~\ref{thm:NFLLMDataSize} suggests that such assumptions are required to prove the success of LMs with practical training data size, as they were in the continuous function approximation setting, e.g., \citep{yun2020transformers,takakura2023approximation,kim2024transformers}.
From a practical perspective, Theorem~\ref{thm:NFLLMDataSize} suggests that we must give up trying to succeed in general settings and actively use properties of natural languages; otherwise, the performance will be at the same level as the trivial Algorithm~\ref{alg:trivial}.
Having said that, since finding mathematically tractable assumptions that natural languages satisfy is hard in general, our Theorem~\ref{thm:negligible} and Proposition~\ref{prp:trivial}, holding under mild assumptions, are still significant as a fundamental guarantee. 
\end{remark}

\section{Regarding the proof of Theorem~\ref{thm:InfiniteHallucination}}
\label{sec:ProofHallucination}

We first give the proof of Theorem~\ref{thm:InfiniteHallucination}, then explain the difference between our proof strategy and the previous work's.

\subsection{Proof of Theorem~\ref{thm:InfiniteHallucination}}
\begin{proof}[Proof of Theorem~\ref{thm:InfiniteHallucination}]
We prove the theorem by constructing a specific $ \AcceptableMap: \SymbolSet^{*} \rightarrow 2^{\SymbolSet^{*}}$.
Since each of $\LMSet$ and $\SymbolSet^{*}$ is a countably infinite set, we can order each of them to obtain an infinite sequence $\LM_{1}, \LM_{2},\dots \in \LMSet$ and $\String_{1}, \String_{2}, \dots \in \SymbolSet^{*}$. For example, we can order them in ascending order with respect to the Godel number.
For $\IDatum \in \Integer_{> 0}$, define $\HallucinationSet_{\IDatum} \subset \Integer_{> 0}$ by
\[
\HallucinationSet_{\IDatum} := \{k \in \Integer_{> 0} \mid \forall \IDatum' = 1, 2, \dots, \IDatum, \String_{k} \neq \LM_{\IDatum'} (\String_{\IDatum})\}.
\]
Here, $\HallucinationSet_{\IDatum}$ is NOT empty since we can construct it by excluding at most finite elements from $\Integer_{>0}$, which is an infinite set.
Hence, we can define $\StringMap: \Integer_{>0} \rightarrow \Integer_{>0}$ by
\[\StringMap (\IDatum) := \min \HallucinationSet_{\IDatum}.\]
Define $\AcceptableFunc: \SymbolSet^{*} \rightarrow \SymbolSet^{*}$ by $\AcceptableFunc (\String_{\IDatum}) := \String_{\StringMap (\IDatum)}$ and $\AcceptableMap$ by $\AcceptableMap (\String) = \{\AcceptableFunc (\String)\}.$
Then, clearly, $|\AcceptableMap (\String)| > 0$, and from the construction of $\StringMap$, for any $k \in \Integer_{> 0}$ the inequality $\LM_{k} (\String_{\IDatum}) \neq \String_{\StringMap (\IDatum)} = \AcceptableFunc (\String_{\IDatum})$ holds. 
In other words, $\LM_{k}$ hallucinates on infinitely many strings $\String_{k}, \String_{k+1}, \String_{k+2}, \dots$, which completes the proof.
\end{proof}

\subsection{Motivation of avoiding depending on the axiom of choice (AC)}

We remark that our proof of Theorem~\ref{thm:InfiniteHallucination} does not use the axiom of choice (AC), while the previous work's proof \cite{xu2024hallucination} depends on the AC.
In this subsection, we discuss its significance.

As an axiomatic system of set theory, most mathematicians use either ZF (Zermelo-Fraenkel set theory) or ZFC, which consists of all the axioms of ZF combined with the axiom of choice (AC), as an axiomatic system of set theory.
We leave the details of the axioms of ZF to textbooks, e.g., \citep{kunen2014set} and below show a version of the AC (Definition 1.1.,  \citep{herrlich2006axiom}).

\begin{definition}
For each family $(A_{i})_{i \in I}$ of non-empty sets $A_{i}$, the product set $\prod_{i \in I} A_{i}$ is non-empty.
\end{definition}

We remark that the statement can be proved from ZF when the index set $I$ is a finite set.
Hence, the difference lies in the cases where the index set $I$ is an infinite set.

At one glance, the statement of the AC should be true, and on ZF, the axiom is equivalent to useful propositions, such as Zorn's lemma, the well-ordering theorem, the existence of basis in every linear space, etc. However, we can also prove some "counterintuitive results," such as the Banach-Tarski paradox from the AC.
For this reason, both ZF and ZFC have been intensively studied in the field of axiomatic set theory, while most fields of mathematics, like algebra and analysis, tend to assume the AC implicitly.

However, our interest is in the physical behavior of computers, not in differences coming from axiomatic systems.
Since the discussion around Theorem~\ref{thm:InfiniteHallucination} is about the physical behavior of computing devices, the results should not depend on the choice of an axiomatic system as long as the axiom system is consistent with physical phenomena. 
In other words, any purely computer-related theorem should be proved, regardless of the choice of a widely used axiom system.
This is why we are interested in avoiding our proof's dependence on the AC.

\subsection{The dependency of the previous work's proof on the AC and how we avoided it}
Now, let us see how the previous work's proof uses the AC and how we have avoided it.
Specifically, when constructing $\StringMap$, the previous work's proof \cite{xu2024hallucination} \Emph{arbitrarily} chose an element from each of $\HallucinationSet_{\IDatum}$ from $\IDatum = 1, 2, \dots$.
Since $\IDatum$ is in the infinite set $\Integer_{> 0}$, such a construction of $\StringMap$ is not guaranteed to exist without the axiom of choice.
Specifically, if $\prod_{\IDatum \in \Integer_{> 0}} \HallucinationSet_{\IDatum}$ is empty, such a $\StringMap$ does not exist.
Hence, we must construct such a function $\StringMap$ or prove $\prod_{\IDatum \in \Integer_{> 0}} \HallucinationSet_{\IDatum}$ in another way.
On the other hand, our proof fixed the order of $\SymbolSet^{*}$ and $\LMSet$ beforehand, and constructed a specific $\StringMap$ using the $\min$ operator.
Hence, our proof is valid even without the axiom of choice.

\subsection{Other parts of this paper and the AC}

We do not investigate dependency on the AC in the other parts of the paper, especially when we consider the probability theory.
As mentioned above, we often assume the AC in many fields of mathematics implicitly, and probability theory is no exception.
This is natural since probability theory allows, e.g., probability mass functions whose value takes non-computable real numbers, which cannot be physically realized by a computer, so such a strong axiom is often necessary to induct results.
Generally speaking, totally excluding the dependency of theories in those areas on the AC is quite demanding and does not make a difference in the implication of the real physical world.
Therefore, we only consider the dependency of the purely computability-theoretic part, i.e., Section~\ref{sec:Innate} on the AC, and we do not make an effort to remove the dependency of the other parts on the AC.

\newpage
\section{Proof of Proposition~\ref{prp:trivial}}
\label{sec:ProofTrivial}

\begin{proof}[Proof of Proposition~\ref{prp:trivial}]
Recall that $\StringLen = \max \{\StringLen' \mid \DataLen > \frac{|\SymbolSet^{*}|^{(\StringLen' + 1)}}{1 - \overline{\CDF} (\StringLen')} \ln \frac{|\SymbolSet^{*}|^{(\StringLen' + 1)}}{2 (1 - \overline{\CDF} (\StringLen'))}\}$.
We prove the following lemma.
\begin{lemma}
\label{lem:FLRM}
Let $\Trainer$ be the FLRM algorithm in Algorithm~\ref{alg:trivial}. Then, $\HallucinationProbability_{\Measure} (\Trainer (\TrainingDataRV)) < \epsilon'_{\HallucinationSymbol}$ holds in probability at least $1 - \epsilon'_{\TrainSymbol}$ over choice of training data sequence $(\StringRV_{1}, \StringRV_{2},\dots,\StringRV_{\DataLen})$, where $\epsilon'_{\HallucinationSymbol} = \epsilon'_{\TrainSymbol} = 2 (1 - \overline{\CDF} (\StringLen))$.
\end{lemma}
Once this lemma is proved, then for any $\epsilon_{\HallucinationSymbol}, \epsilon_{\TrainSymbol} \in (0, 1)$, we obtain that $\HallucinationProbability_{\Measure} (\Trainer (\TrainingDataRV)) < \epsilon_{\HallucinationSymbol}$ holds in probability at least $1 - \epsilon_{\TrainSymbol}$ over choice of training data sequence if $\DataLen > \frac{|\SymbolSet^{*}|^{(\bar{\StringLen} + 1)}}{1 - \overline{\CDF} (\bar{\StringLen})} \ln \frac{|\SymbolSet^{*}|^{(\bar{\StringLen} + 1)}}{2(1 - \overline{\CDF} (\bar{\StringLen}))}$, where $\bar{\StringLen}$ is an integer that satisfies $1 - \overline{\CDF} (\bar{\StringLen}) < \frac{1}{2} \min \{\epsilon_{\HallucinationSymbol}, \epsilon_{\TrainSymbol}\}$, which completes the proof of Proposition~\ref{prp:trivial}.
Note that such a $\bar{\StringLen}$ exists since $\overline{\CDF}$ is non-decreasing and $\lim_{\StringLen' \to +\infty} \overline{\CDF} (\StringLen') = 1$ by assumptions.

Now, we prove Lemma~\ref{lem:FLRM}. 
We first decompose the hallucination probability as follows: 
\begin{align}
\HallucinationProbability_{\Measure} (\Trainer (\TrainingDataRV)) &:= \Probability_{\StringRV \sim \Measure} (\Trainer (\TrainingDataRV) (\StringRV) \notin \AcceptableMap (\StringRV)) \nonumber\\ 
&= \Probability_{\StringRV \sim \Measure} (\Trainer (\TrainingDataRV) (\StringRV) \notin \AcceptableMap (\StringRV) \text{ and } \Length(\StringRV) \le \StringLen) \nonumber \\ 
& \quad + \Probability_{\StringRV \sim \Measure} (\Trainer (\TrainingDataRV) (\StringRV) \notin \AcceptableMap (\StringRV) \text{ and } \Length(\StringRV) > \StringLen) \nonumber\\
&\le \Probability_{\StringRV \sim \Measure} (\Trainer (\TrainingDataRV) (\StringRV) \notin \AcceptableMap (\StringRV) \text{ and } \Length(\StringRV) \le \StringLen) + \Probability_{\StringRV \sim \Measure} (\Length(\StringRV) > \StringLen) \nonumber\\
&\le \Probability_{\StringRV \sim \Measure} (\Trainer (\TrainingDataRV) (\StringRV) \notin \AcceptableMap (\StringRV) \text{ and } \Length(\StringRV) \le \StringLen) + \epsilon'_{\HallucinationSymbol}/2.
\label{eqn:HallucinationDecomposition}
\end{align}
Here, $\Probability (\Length (\StringRV) > \StringLen) \le \epsilon'_{\HallucinationSymbol}/2$ is due to the definition of $\overline{\CDF}$ and $\epsilon'_{\HallucinationSymbol} = 2(1 - \overline{\CDF} (\StringLen))$.
Remark that if $\Probability_{\StringRV \sim \Measure} (\Trainer (\TrainingDataRV) (\StringRV) \notin \AcceptableMap (\StringRV) \text{ and } \Length(\StringRV) \le \StringLen) < \epsilon'_{\HallucinationSymbol}/2$, then $\HallucinationProbability_{\Measure} (\Trainer (\TrainingDataRV)) < \epsilon'_{\HallucinationSymbol}$ holds by \eqref{eqn:HallucinationDecomposition}.
In the following, we evaluate
$\Probability_{\StringRV \sim \Measure} (\Trainer (\TrainingDataRV) (\StringRV) \notin \AcceptableMap (\StringRV) \text{ and } \Length(\StringRV) \le \StringLen)$.
Denote the set of strings no longer than $\StringLen$ by $\SymbolSet^{(\le \StringLen)}$, defined by $\SymbolSet^{(\le \StringLen)} := \SymbolSet^{0} \cup \SymbolSet^{1} \cup \dots \cup \SymbolSet^{\StringLen}$.
Also, define $\NStrings := |\SymbolSet^{(\le \StringLen)}| \in \Integer_{\ge 0}$.
Note that $\NStrings := |\SymbolSet^{(\le \StringLen)}| = |\SymbolSet^{0} \cup \SymbolSet^{1} \cup \dots \cup \SymbolSet^{\StringLen}| = \frac{|\SymbolSet|^{\StringLen} - 1}{|\SymbolSet| - 1} \le |\SymbolSet|^{\StringLen}$ holds.
We index all the elements in $\SymbolSet^{(\le \StringLen)}$ in the descending order with respect to its probability. 
In other words, $\String_{1}, \String_{2}, \dots, \String_{\NStrings} \in \SymbolSet^{(\le \StringLen)}$ satisfy $\String_{\IString} \neq \String_{\IString'}$ and $\Probability (\StringRV = \String_{\IString}) \ge \Probability (\StringRV = \String_{\IString'})$ for any $\IString, \IString'$ satisfying $1 \le \IString < \IString' \le \NStrings$ and any random variable $\StringRV$ generated by the distribution $\Measure$.
For $\IString = 1, 2, \dots, \NStrings$, define $\PrMass_{\IString} = \Probability (\StringRV = \String_{\IString})$.
Also, define $\IString^{(*)} := \min \{\IString \in \{1, 2, \dots, \NStrings\} \mid \sum_{\IString'=\IString + 1}^{\NStrings} \PrMass_{\IString'} < (\epsilon'_{(\HallucinationSymbol)})/2\}$.
In other words, $\IString^{(*)}$ is the unique index that satisfies $\sum_{\IString'=\IString^{(*)}}^{\NStrings} \PrMass_{\IString'} \ge (\epsilon'_{(\HallucinationSymbol)})/2$ and $\sum_{\IString'=\IString^{(*)}+1}^{\NStrings} \PrMass_{\IString'} < (\epsilon'_{(\HallucinationSymbol)})/2$.
Let $\StringRV_{1}, \StringRV_{2}, \dots, \StringRV_{\DataLen}$ be mutually independent random variables, all generated by the distribution $\Measure$.
Here, if $\{\StringRV_{1}, \StringRV_{2}, \dots, \StringRV_{\DataLen}\} \supset \{\String_{1}, \String_{2},\dots, \String_{\IString^{(*)}}\}$ holds, then the output of Algorithm~\ref{alg:trivial} does not hallucinates on any inputs in $\{\String_{1}, \String_{2},\dots, \String_{\IString^{(*)}}\}$, which leads to 
$\Probability_{\StringRV \sim \Measure} (\Trainer (\TrainingDataRV) (\StringRV) \notin \AcceptableMap (\StringRV) \text{ and } \Length(\StringRV) \le \StringLen) < \epsilon'_{\HallucinationSymbol}/2$.
Hence, all we need to do is to upper bound the probability of the event $\{\StringRV_{1}, \StringRV_{2}, \dots, \StringRV_{\DataLen}\} \not\supset \{\String_{1}, \String_{2},\dots, \String_{\IString^{(*)}}\}$. 
We denote this event by $\HallucinationEvent_{(\le \IString^{(*)})}^{(\DataLen)}$.
In the following, we evaluate $\Probability (\HallucinationEvent_{(\le \IString^{(*)})}^{(\DataLen)})$. 

Let $\HallucinationEvent_{(\IString)}^{(\DataLen)}$ denote the event $(\StringRV_{1} \neq \String_{\IString}) \text{ and } (\StringRV_{2} \neq \String_{\IString}) \text{ and } \dots \text{ and } (\StringRV_{\DataLen} \neq \String_{\IString})$.
Since $\HallucinationEvent_{(\le \IString^{(*)})}^{(\DataLen)} = \bigcup_{\IString=1}^{\IString^{(*)}} \HallucinationEvent_{(\IString)}^{(\DataLen)}$ holds, we have that $\Probability (\HallucinationEvent_{(\le \IString^{(*)})}^{(\DataLen)}) \le \sum_{\IString=1}^{\IString^{(*)}} \Probability (\HallucinationEvent_{(\IString)}^{(\DataLen)})$.
Recall that $\PrMass_{\IString} := \Probability (\StringRV_{\IDatum} = \String_{\IString})$ for all $\IDatum = 1, 2, \dots, \DataLen$.
Since $\StringRV_{1}, \StringRV_{2}, \dots, \StringRV_{\DataLen}$ are mutually independent, we have that $\Probability (\HallucinationEvent_{(\IString)}^{(\DataLen)}) = (1 - \PrMass_{\IString})^{\DataLen}$.
By the definition of the indexing order, $\PrMass_{\IString} \le \PrMass_{\IString^{(*)}}$ holds for $\IString = \IString^{(*)}, \IString^{(*)} + 1,\dots, \NStrings$.
Also, by the definition of $\IString^{(*)}$, we have that $\sum_{\IString = \IString^{(*)}}^{\NStrings} \PrMass_{\IString} \ge \epsilon'_{\HallucinationSymbol}/2$.
Therefore, we obtain
\[
\PrMass_{\IString^{(*)}} = \max \{\PrMass_{\IString^{(*)}}, \PrMass_{\IString^{(*)} + 1}, \dots, \PrMass_{\NStrings}\} \ge \frac{\sum_{\IString = \IString^{(*)}}^{\NStrings} \PrMass_{\IString}}{\NStrings - \IString^{(*)} + 1} \ge \frac{\epsilon'_{\HallucinationSymbol}}{2 (\NStrings - \IString^{(*)} + 1)} \ge \frac{\epsilon'_{\HallucinationSymbol}}{2 \NStrings}.
\]
Here, the second inequality comes from the fact that the maximum value is always larger than or equal to the mean.
Thus, for $\IString = 1, 2, \dots, \IString^{(*)}$, we have that $\PrMass_{\IString} \ge \frac{\epsilon_{\HallucinationSymbol}}{2 \NStrings}$.
Hence, we obtain $\Probability (\HallucinationEvent_{(\IString)}^{(\DataLen)}) = (1 - \PrMass_{\IString})^{\DataLen} \le (1 - \frac{\epsilon'_{\HallucinationSymbol}}{2 \NStrings})^{\DataLen} \le \exp(-\frac{\DataLen \epsilon'_{\HallucinationSymbol}}{2 \NStrings})$, where the last inequality comes from the general inequality $1 + x \le \exp(x)$.
Therefore, 
\begin{align*}
\Probability (\HallucinationEvent_{(\le \IString^{(*)})}^{(\DataLen)}) 
&\le \sum_{\IString=1}^{\IString^{(*)}} \Probability (\HallucinationEvent_{(\IString)}^{(\DataLen)}) 
\le \sum_{\IString=1}^{\IString^{(*)}} \exp\left(-\frac{\DataLen \epsilon'_{\HallucinationSymbol}}{2 \NStrings}\right) \\
&\le \IString^{(*)} \exp\left(-\frac{\DataLen \epsilon'_{\HallucinationSymbol}}{2 \NStrings}\right) 
\le \NStrings \exp\left(-\frac{\DataLen \epsilon'_{\HallucinationSymbol}}{2 \NStrings}\right) \\ 
&\le |\SymbolSet|^{\StringLen + 1} \exp\left(-\frac{\DataLen \epsilon'_{\HallucinationSymbol}}{2 |\SymbolSet|^{\StringLen + 1}}\right).
\end{align*}
From the above, we can see that $\DataLen > \frac{2 |\SymbolSet|^{\StringLen + 1}}{\epsilon'_{\HallucinationSymbol}} \ln \frac{|\SymbolSet|^{\StringLen + 1}}{\epsilon'_{\TrainSymbol}}$ is a sufficient condition for $\Probability (\HallucinationEvent_{(\le \IString^{(*)})}^{(\DataLen)}) < \epsilon'_{\TrainSymbol}$ to hold.
This completes the proof of Lemma~\ref{lem:FLRM}.
\end{proof}

\newpage
\section{Proof strategy for LM limitations and no-free-lunch-theorem}
\label{sec:ProofNegative}

Proposition~\ref{prp:trivial}, essentially obtained by Proposition~\ref{prp:trivial}, was proved by a specific algorithm.
To prove such a positive result, constructing an algorithm suffices.
On the other hand, to prove negative results such as Theorem~\ref{thm:NFLLM} and Theorem~\ref{thm:NFLLMDataSize}, we need to show that the negative events happen no matter what algorithm we use.
Thus, a simple construction-based proof does not work, and instead, we rely on the no-free-lunch theorem.

Indeed, we can prove Theorem~\ref{thm:NFLLM} and Theorem~\ref{thm:NFLLMDataSize} by the following no free lunch theorem proved in \citep{shalev2014understanding}.

\begin{theorem}[General no free lunch theorem]
\label{thm:NFL}
Consider a learning problem from a domain set $\Domain$ to a codomain set $\Codomain$ such that $|\Codomain| \ge 1$, i.e., $\Codomain \neq \emptyset$.
For a probability measure $\Measure$ on $\Domain$, a ground truth map $\AcceptableFunc: \Domain \rightarrow \Codomain$, denote the hallucination probability (0-1 risk) of a hypothesis map $\Hypothesis: \Domain \rightarrow \Codomain$ on $\Measure$ and $\AcceptableFunc$ by $\HallucinationProbability_{\Measure, \AcceptableFunc} (\Hypothesis)$, which is defined by $\HallucinationProbability_{\Measure, \AcceptableFunc} (\Hypothesis) = \Probability (\Hypothesis(\InputRV) \neq \AcceptableFunc (\InputRV))$.
Then, for any map (learning algorithm) $\Learner: (\Domain \times \Codomain)^{*} \rightarrow (\Domain \rightarrow \Codomain)$, any nonnegative integer (training data size) $\DataLen$ that satisfies $\DataLen \le \frac{1}{2} |\Domain|$, any finite positive integer $\NLabels$ satisfying $1 \le \NLabels \le |\Codomain|$, and any $\LowerBound_{\HallucinationSymbol} \in (0, 1)$, there exist a computable map $\AcceptableFunc: \Domain \rightarrow \Codomain$ and a finite subset $\underlineDomain \subset \Domain$ such that both the following inequalities hold.
\begin{align*}
\Expect_{\TrainingDataRV} \HallucinationProbability_{\Uniform (\underlineDomain), \AcceptableFunc} (\Learner (\TrainingDataRV)) &\ge \HPExpect := \frac{\NLabels-1}{2 \NLabels}, \\
\Probability_{\TrainingDataRV} (\HallucinationProbability_{\Uniform (\underlineDomain), \AcceptableFunc} (\Learner (\TrainingDataRV)) \ge \LowerBound_{\HallucinationSymbol}) &\ge \LowerBound_{(\TrainSymbol, \NLabels)} := \frac{\HPExpect - \LowerBound_{\HallucinationSymbol}}{1 - \LowerBound_{\HallucinationSymbol}} = \frac{\NLabels - 1 - 2 \NLabels  \LowerBound_{\HallucinationSymbol}}{2 \NLabels - 2 \NLabels \LowerBound_{\HallucinationSymbol}}.
\end{align*}
Here, $\TrainingDataRV = ((\underline{\InputRV}_{1}, \OutputRV_{1}), (\underline{\InputRV}_{2}, \OutputRV_{2}), \dots, (\underline{\InputRV}_{\DataLen}, \OutputRV_{\DataLen}))$ is a length-$\DataLen$ random training data sequence, where $\underline{\InputRV}_{1}, \underline{\InputRV}_{2}, \dots, \underline{\InputRV}_{\DataLen}$ are i.i.d. random variables generated by $\Uniform (\underlineDomain)$, the uniform distribution on $\underlineDomain$, and $\OutputRV_{\IDatum} = \AcceptableFunc (\underline{\InputRV}_{\IDatum})$ for $\IDatum = 1, 2, \dots, \DataLen$
and the operators $\Expect_{\TrainingDataRV}$ and $\Probability_{\TrainingDataRV}$ return the expectation of the return value of the given function of the random variable $\TrainingDataRV$ and the probability of the given condition depending on $\TrainingDataRV$ being satisfied, respectively.
\end{theorem}

\begin{remark}
We are interested in the cases where $|\Codomain| \ge 2$ and we can take $\NLabels$ so that $\NLabels \ge 2$.
If $\NLabels \ge 2$, then $\HPExpect \ge 1/4$ and $\LowerBound_{\TrainSymbol} \ge \frac{1 - 4 \LowerBound_{\HallucinationSymbol}}{4 - 4 \LowerBound_{\HallucinationSymbol}}$.
Moreover, if $\LowerBound_{\HallucinationSymbol} = 1/8$, then $\LowerBound_{\TrainSymbol} \ge 1/7$.
\end{remark}

Theorem~\ref{thm:NFL} holds even where $\Domain$ or $\Codomain$ is infinite.
If $\Codomain$ is an infinite set, the following corollary is useful.

\begin{corollary}
\label{cor:InfiniteCodomainNFL}
Under the setting of Theorem~\ref{thm:NFL}, if $\Codomain$ is an infinite set, then for any map (learning algorithm) $\Learner: (\Domain \times \Codomain)^{*} \rightarrow (\Domain \rightarrow \Codomain)$, any nonnegative integer (training data size) $\DataLen$ that satisfies $\DataLen \le \frac{1}{2} |\Domain|$, and any $\LowerBound_{\HallucinationSymbol} \in (0, 1)$,
there exist a computable map $\AcceptableFunc: \Domain \rightarrow \Codomain$ and a finite subset $\underlineDomain \subset \Domain$ such that both the following inequalities hold:
\begin{align*}
\Expect_{\TrainingDataRV} \HallucinationProbability_{\Uniform (\underlineDomain), \AcceptableFunc} (\Learner (\TrainingDataRV)) &\ge 1/2, \\
\Probability_{\TrainingDataRV} (\HallucinationProbability_{\Uniform (\underlineDomain), \AcceptableFunc} (\Learner (\TrainingDataRV)) \ge \LowerBound_{\HallucinationSymbol}) &\ge \LowerBound_{\TrainSymbol} := \frac{1 - 2 \LowerBound_{\HallucinationSymbol}}{2 - 2 \LowerBound_{\HallucinationSymbol}}.
\end{align*}
\end{corollary}

\begin{proof}
For any $\epsilon \in \Real_{>0}$, we can prove that $\Expect_{\TrainingDataRV} \HallucinationProbability_{\Uniform (\underlineDomain), \AcceptableFunc} (\Learner (\TrainingDataRV)) \ge 1/2 - \epsilon$ by taking a sufficiently large $\NLabels$ in Theorem~\ref{thm:NFL}.
Hence, $\Expect_{\TrainingDataRV} \HallucinationProbability_{\Uniform (\underlineDomain), \AcceptableFunc} (\Learner (\TrainingDataRV)) < 1/2$ cannot hold, which completes the proof of $\Expect_{\TrainingDataRV} \HallucinationProbability_{\Uniform (\underlineDomain), \AcceptableFunc} (\Learner (\TrainingDataRV)) \ge 1/2$.
We can prove $\Probability_{\TrainingDataRV} (\HallucinationProbability_{\Uniform (\underlineDomain), \AcceptableFunc} (\Learner (\TrainingDataRV)) \ge \LowerBound_{\HallucinationSymbol}) \ge \LowerBound_{\TrainSymbol}$ by applying Lemma~\ref{lem:MarkovVariant} with $c=1$ and $a = \LowerBound_{\HallucinationSymbol}$ to $\Expect_{\TrainingDataRV} \HallucinationProbability_{\Uniform (\underlineDomain), \AcceptableFunc} (\Learner (\TrainingDataRV)) \ge 1/2$.
\end{proof}

\begin{remark}
In Corollary~\ref{cor:InfiniteCodomainNFL}, for example, if $\LowerBound_{\HallucinationSymbol} = 1/4$, then $\LowerBound_{\TrainSymbol} = 1/3$.
\end{remark}

\begin{remark}[Regarding the statement of Theorem~\ref{thm:NFL}]
\label{rmk:NFL}
Theorem~\ref{thm:NFL} is a generalized version of the no free lunch theorem given as \citep[Theorem 5.1.]{shalev2014understanding} as Theorem~\ref{thm:NFL} provides a tighter bound when $|\Codomain| \ge 3$. However, the proof technique is essentially the same as that of Theorem 5.1. in \citep{shalev2014understanding}.
We also remark that the computability of the map $\AcceptableFunc$ was pointed out by \citep{agarwal2020learnability}.
\end{remark}

From Theorem~\ref{thm:NFL}, we obtain Theorem~\ref{thm:NFLLM} and Theorem~\ref{thm:NFLLMDataSize} as follows.

\begin{proof}[Proof of Theorem~\ref{thm:NFLLM}]
For any $\DataLen$, consider a subset $\Domain \subset \SymbolSet^{*}$ such that $|\Domain| > 2 \DataLen$.
We obtain the theorem by applying Theorem~\ref{thm:NFL} to $\Domain$.
\end{proof}

\begin{proof}[Proof of Theorem~\ref{thm:NFLLMDataSize}]
First, the following lemma holds, whose proof is in Appendix.
\begin{lemma}
\label{lem:SetsFromCDF}
We can construct a set $\Domain \subset \SymbolSet^{*}$ such that $|\Domain| = \lfloor\frac{|\SymbolSet|^{\underline{\StringLen} + 1} - 1}{(|\SymbolSet| - 1) \overline{\CDF} (\underline{\StringLen})}\rfloor$ and $\CDF_{\Length \sharp \Uniform (\Domain)} (\StringLen) \ge \overline{\CDF} (\StringLen)$ for all $\StringLen \in \Integer_{\ge 0}$.
\end{lemma}
Once we admit Lemma~\ref{lem:SetsFromCDF}, then we obtain Theorem~\ref{thm:NFLLMDataSize} by applying Theorem~\ref{thm:NFL} to the set $\Domain$ constructed by Lemma~\ref{lem:SetsFromCDF}, which completes the proof.
\end{proof}

The proof of Lemma~\ref{lem:SetsFromCDF} is given as follows.

\begin{proof}[Proof of Lemma~\ref{lem:SetsFromCDF}]
Recall that $\underline{\StringLen} := \argmin_{\StringLen \in \Integer_{\ge 0}} \frac{|\SymbolSet|^{\StringLen + 1} - 1}{(|\SymbolSet| - 1) \overline{\CDF} (\StringLen)}$.
For $\StringLen \in \Integer_{\ge 0}$, construct $\Domain_{\StringLen}$ as follows.
\begin{itemize}
    \item If $\frac{|\SymbolSet|^{\StringLen + 1} - 1}{|\SymbolSet| - 1} \le \frac{|\SymbolSet|^{\underline{\StringLen} + 1} - 1}{(|\SymbolSet| - 1) \overline{\CDF} (\underline{\StringLen})}$, then $\Domain_{\StringLen} := \SymbolSet^{\StringLen}$,
    \item If $\frac{|\SymbolSet|^{\StringLen} - 1}{|\SymbolSet| - 1} \le \frac{|\SymbolSet|^{\underline{\StringLen} + 1} - 1}{(|\SymbolSet| - 1) \overline{\CDF} (\underline{\StringLen})} < \frac{|\SymbolSet|^{\StringLen + 1} - 1}{|\SymbolSet| - 1}$, then construct $\Domain_{\StringLen}$ by collecting arbitrary $\lfloor\frac{|\SymbolSet|^{\underline{\StringLen} + 1} - 1}{(|\SymbolSet| - 1) \overline{\CDF} (\underline{\StringLen})}\rfloor - \frac{|\SymbolSet|^{\StringLen} - 1}{|\SymbolSet| - 1}$ elements in $\SymbolSet^{\StringLen}$,
    \item If $\frac{|\SymbolSet|^{\underline{\StringLen} + 1} - 1}{(|\SymbolSet| - 1) \overline{\CDF} (\underline{\StringLen})} < \frac{|\SymbolSet|^{\StringLen} - 1}{|\SymbolSet| - 1}$, then $\Domain_{\StringLen} = \emptyset$.
\end{itemize}
Then, the set $\Domain := \Domain_{0} \cup \Domain_{1} \cup \dots$ satisfies the following:
\begin{enumerate}
    \item $|\Domain| = \lfloor\frac{|\SymbolSet|^{\underline{\StringLen} + 1} - 1}{(|\SymbolSet| - 1) \overline{\CDF} (\underline{\StringLen})}\rfloor$, and
    \item Let $\InputRV$ be a random variable generated by $\Uniform (\Domain)$. For any $\StringLen \in \Integer_{\ge 0}$, $\Probability (\Length(\InputRV) \le \StringLen) = |\Domain_{(\le \StringLen)}|/|\Domain| \ge \overline{\CDF} (\StringLen)$, where $\Domain_{(\le \StringLen)} := \Domain_{0} \cup \Domain_{1} \cup \dots \cup \Domain_{\StringLen}$,
\end{enumerate}
which are the consequences of Lemma~\ref{lem:FLRM}.
Hence, we can complete the proof by confirming the above two properties.
Here, the first property is trivial.
Noting that $|\Domain| \le \frac{|\SymbolSet|^{\underline{\StringLen} + 1} - 1}{(|\SymbolSet| - 1) \overline{\CDF} (\underline{\StringLen})}$, the second property can be confirmed as follows.
\begin{itemize}
    \item If $\frac{|\SymbolSet|^{\StringLen + 1} - 1}{|\SymbolSet| - 1} \le \frac{|\SymbolSet|^{\underline{\StringLen} + 1} - 1}{(|\SymbolSet| - 1) \overline{\CDF} (\underline{\StringLen})}$, then $\Domain_{(\le \StringLen)} = \SymbolSet^{(\le \StringLen)}$, so $|\Domain_{(\le \StringLen)}| = \frac{|\SymbolSet|^{\StringLen + 1} - 1}{|\SymbolSet| - 1}$.
    Hence $|\Domain_{(\le \StringLen)}|/|\Domain| \ge \frac{|\SymbolSet|^{\StringLen + 1} - 1}{|\SymbolSet| - 1} \cdot \frac{(|\SymbolSet| - 1) \overline{\CDF} (\underline{\StringLen})}{|\SymbolSet|^{\underline{\StringLen} + 1} - 1} \ge \frac{|\SymbolSet|^{\StringLen + 1} - 1}{|\SymbolSet| - 1} \cdot \frac{(|\SymbolSet| - 1) \overline{\CDF} (\StringLen)}{|\SymbolSet|^{\StringLen + 1} - 1} = \overline{\CDF} (\StringLen)$ holds, where the second inequality is due to the definition of $\underline{\StringLen}$.
    \item If $\frac{|\SymbolSet|^{\underline{\StringLen} + 1} - 1}{(|\SymbolSet| - 1) \overline{\CDF} (\underline{\StringLen})} < \frac{|\SymbolSet|^{\StringLen + 1} - 1}{|\SymbolSet| - 1}$, then since $\Domain_{(\le \StringLen)} = \Domain$, the inequality $|\Domain_{(\le \StringLen)}|/|\Domain| = 1 \ge \overline{\CDF} (\StringLen)$ is trivial since $\overline{\CDF} (\StringLen) \in [0, 1]$ by definition.
\end{itemize}
These complete the proof.
\end{proof}

We conclude this section with a complete proof of Theorem~\ref{thm:NFL}.
\begin{proof}[Proof of Theorem~\ref{thm:NFL}]
The statement is trivial if $\NLabels = 1$. Hence, in the following, we assume that $|\Codomain|>1$ and $1 \le \NLabels \le |\Codomain|$.
Recall that for any positive integer $a$, we denote the set $\{1, 2, \dots, a\}$ by $[a]$.
Recall that $\DataLen \le \frac{1}{2} |\Domain|$.
Define $\underlineDomain := \Domain$ if $\Domain$ is a finite set and let $\underlineDomain$ be an arbitrary finite subset of $\Domain$ satisfying $|\underlineDomain| = 2 \DataLen$ if $\Domain$ is an infinite set.
In any case, $\NInputs := |\underlineDomain| \ge 2 \DataLen$ is satisfied.
Likewise, recall that $\NLabels \le |\Codomain|$ and let $\underlineCodomain$ be a finite subset of $\Codomain$ satisfying $|\underlineCodomain| = \NLabels$.
Let $\Codomain^{\underlineDomain}$ denote the set of all the maps from $\underlineDomain$ to $\Codomain$.
Likewise, let $\underlineCodomain^{\underlineDomain}$ denote the set of all the maps from $\underlineDomain$ to $\underlineCodomain$.
Clearly, $|\underlineCodomain^{\underlineDomain}| = |\underlineCodomain|^{|\underlineDomain|} = \NLabels^{\NInputs}$. 
In other words, there are $\NFuncs := \NLabels^{\NInputs}$ possible maps from $\underlineDomain$ to $\underlineCodomain$.
We index these maps so that we have a sequence $\Func_{1}, \Func_{2}, \dots, \Func_{\NFuncs}$ of distinct maps such that $\{\Func_{1}, \Func_{2}, \dots, \Func_{\NFuncs}\} = \underlineCodomain^{\underlineDomain}$.
In the following, for $\InputSeq = (\Input_{1}, \Input_{2}, \dots, \Input_{\DataLen}) \in \Domain^{\DataLen}$ and $\Func \in \Codomain^\Domain$, let $\Func (\InputSeq)$ denote $(\Func (\Input_{1}), \Func (\Input_{2}), \dots, \Func (\Input_{\DataLen})) \in \Codomain^{\DataLen}$ and $(\InputSeq, \Func(\InputSeq))^\top$ denote $((\Input_{1}, \Func (\Input_{1})), (\Input_{2}, \Func (\Input_{2})), \dots, (\Input_{\DataLen}, \Func (\Input_{\DataLen}))) \in (\Domain \times \Codomain)^{\DataLen}$. 
We are to prove that for any map $\Trainer: \Domain^{\DataLen} \rightarrow \Codomain^{\Domain}$, the following holds:
\begin{equation}
\max_{\IFunc \in [\NFuncs]} \Expect_{\InputSeqRV \sim \Uniform (\underlineDomain)^{\DataLen}} \HallucinationProbability_{\Uniform (\underlineDomain), \Func_{\IFunc}} (\Trainer ((\InputSeqRV, \Func_{\IFunc} (\InputSeqRV))^\top)) \ge \frac{\NLabels-1}{2\NLabels},
\label{eqn:NFL-proof1}
\end{equation}
where $\Expect_{\InputSeqRV \sim \Uniform (\underlineDomain)^{\DataLen}}$ indicates the expectation operator with respect to the random variable sequence $\InputSeqRV = (\InputRV_{1}, \InputRV_{2}, \dots, \InputRV_{\DataLen})$, a sequence of independent random variables, each of which is generated by the identical uniform distribution $\Uniform(\underlineDomain)$.
Note that since $\underlineDomain$ and $\underlineCodomain$ are finite sets, $\Func_{\IFunc}$ is a computable map for all $\IFunc \in [\NFuncs]$.
There are $\NDatasets := |\underlineDomain|^\DataLen$ possible data sequences in $\underlineDomain^\DataLen$. 
We index these data sequences so that we have a sequence $\InputSeq_{1}, \InputSeq_{2}, \dots, \InputSeq_{\NDatasets}$ of distinct data sequences satisfying $\{\InputSeq_{1}, \InputSeq_{2}, \dots, \InputSeq_{\NDatasets}\} = \underlineDomain^\DataLen$.
By the definition of the uniform distribution $\Uniform(\underlineDomain)^{\DataLen}$, we have
\begin{equation}
\Expect_{\InputSeqRV \sim \Uniform (\underlineDomain)^{\DataLen}} \HallucinationProbability_{\dots} (\Trainer (\dots)) = \frac{1}{\NDatasets} \sum_{\IDataset=1}^{\NDatasets} \HallucinationProbability_{\dots} (\Trainer ((\InputSeq_{\IDataset}, \Func_{\IFunc} (\InputSeq_{\IDataset}))^\top)).
\label{eqn:NFL-proof2}
\end{equation}
The rest of the proof follows the standard no-free-lunch theorem argument. We omit the details which can be found in, e.g., \citep{shalev2014understanding}.
The key idea is to average the risk over all possible functions and datasets, showing that for any learner, there must be a function for which it performs poorly. This leads to the desired lower bound on the expected risk.

The second part of the theorem, concerning the probability of high risk, is a direct consequence of the first part combined with the following variant of Markov's inequality.
\begin{lemma}
\label{lem:MarkovVariant}
Let $c$ be a positive real number and let $Z$ be a random variable taking a value in $[0, c]$ and assume its expectation is given by $\Expect Z = \mu \in \Real$. Then, for any real number $a \in (0, c)$, the following inequality holds:
\[
\Probability(Z > a) \ge \frac{\mu - a}{c - a}.
\]
\end{lemma}
\begin{proof}
Noting that $Z \le a$ is equivalent to $c - Z \ge c - a$ and that $c - Z$ is a nonnegative random variable whose expectation is $c - \mu$, we obtain the following evaluation by Markov's inequality:
\[
\Probability(Z > a) = 1 - \Probability(Z \le a) = 1 - \Probability(c - Z \ge c - a)
\ge 1 - \frac{c - \mu}{c - a} = \frac{\mu - a}{c - a},
\]
where we applied Markov's inequality to the nonnegative random variable $c - Z$ to get the inequality.
\end{proof}
We complete the proof of Theorem~\ref{thm:NFL} by applying Lemma~\ref{lem:MarkovVariant} with $c=1$ and $a = \LowerBound_{\HallucinationSymbol}$ to the expectation result.
\end{proof}

\bibliography{main}

\end{document}